\documentclass[journal]{IEEEtran}

\usepackage{cite}
\usepackage{graphicx}
\usepackage{amsmath}
\usepackage{algorithm}
\usepackage{algorithmic}
\usepackage{array}
\usepackage[caption=false,font=footnotesize]{subfig}
\usepackage{dblfloatfix}
\usepackage{url}
\usepackage{ulem}
\usepackage{caption}
\usepackage{multirow}

\begin{document}

\title{Progressive Multi-Stage Learning for Discriminative Tracking}
%
%
%

\author{Weichao~Li,
        Xi~Li,
        Omar~Elfarouk~Bourahla,
        Fuxian~Huang,
        Fei~Wu,
        Wei~Liu,
        Zhiheng~Wang,
        and~Hongmin~Liu
\thanks{Manuscript received August 22, 2019; revised February 1, 2020; accepted March 16, 2020. Date of publication XXX, 2020; date of current version XXX, 2020.
This work is in part supported by key scientific technological innovation research project by Ministry of Education, Zhejiang Provincial Natural Science Foundation of China under
Grant LR19F020004, Baidu AI Frontier Technology Joint Research Program, Tencent AI Lab Rhino-Bird Joint Research Program(No. JR201806), Zhejiang University K.P.Chao's High Technology Development Foundation.
\textit{(Corresponding author: Xi Li and Hongmin Liu.)}}
\thanks{W. Li, X. Li, O. Bourahla, F. Huang, and F. Wu are with College of Computer Science, Zhejiang University, Hangzhou, China
 (email: $\{$weichaoli, xilizju$\}$@zju.edu.cn, obourahla@ymail.com, hfuxian@zju.edu.cn, wufei@cs.zju.edu.cn).}
\thanks{W. Liu are with AI Lab of Tencent Inc, Shenzhen, China
 (email: wliu@ee.columbia.edu).}%
\thanks{Z. Wang and H. Liu are with College of Computer Science and Technology, Henan Polytechnic University, Jiaozuo, China
 (email: wzhenry@eyou.com, hmliu\_82@163.com).}
 }

\markboth{IEEE Transactions on Cybernetics,~Vol.~XX, No.~X, April~2020}%
{Li \MakeLowercase{\textit{et al.}}: Progressive Multi-Stage Learning for Discriminative Tracking}


\maketitle

\begin{abstract}
Visual tracking is typically solved as a discriminative learning problem that usually requires high-quality samples for online model adaptation. It is a critical and challenging problem to evaluate the training samples collected from previous predictions and employ sample selection by their quality to train the model.

To tackle the above problem, we propose a joint discriminative learning scheme with the progressive multi-stage optimization policy of sample selection for robust visual tracking. The proposed scheme presents a novel time-weighted and detection-guided self-paced learning strategy for easy-to-hard sample selection, which is capable of tolerating relatively large intra-class variations while maintaining inter-class separability. Such a self-paced learning strategy is jointly optimized in conjunction with the discriminative tracking process, resulting in robust tracking results. Experiments on the benchmark datasets demonstrate the effectiveness of the proposed learning framework.
\end{abstract}


%
\IEEEpeerreviewmaketitle

\section{Introduction}

\IEEEPARstart{A}{s} an important and challenging problem in computer vision, visual tracking has a wide range of applications such as visual surveillance, human computer interaction, and video compression. In principle, the goal of visual tracking is to estimate the trajectory of a generic target object in an image sequence, given only its initial state. In practice, the tracking problem is particularly challenging due to the lack of sufficient a-priori knowledge for the target object. Moreover, the target object has to undergo various unpredictable transformations caused by several complicated factors (e.g., illumination variation, partial occlusion, and shape deformation), which are usually likely to contaminate the tracking process with error accumulation.

\begin{figure*}[!htbp]
\centering
\includegraphics[width=0.8\textwidth]{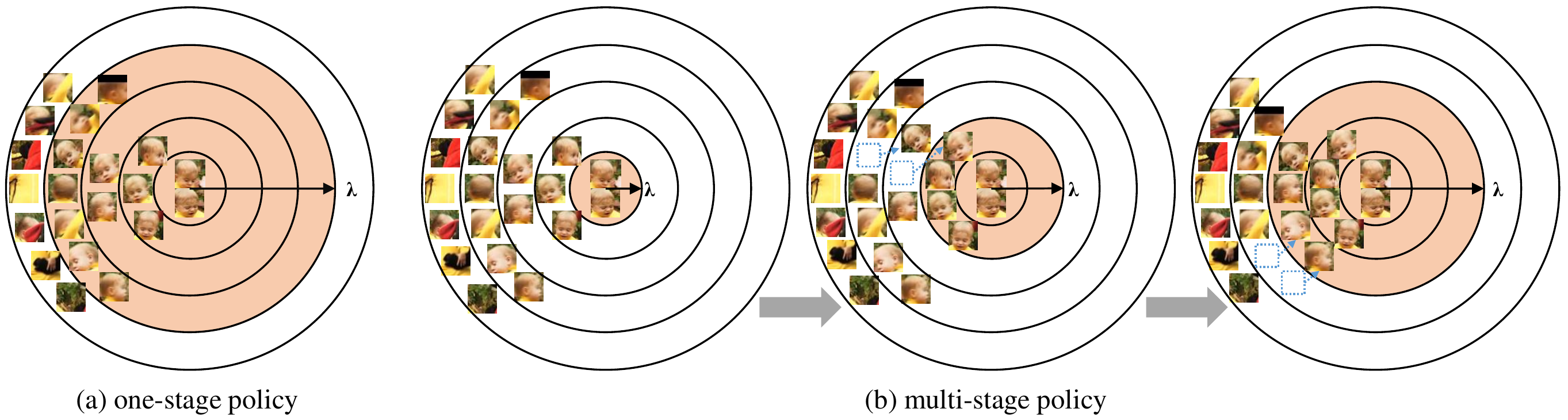}
\caption{A schematic illustration of the differences between the one-stage and the multi-stage policy of sample selection ($\lambda$ denotes the learning pace and the selected samples are within the red circle). Compared to the one-stage policy, our multi-stage policy progressively performs sample selection in an easy-to-hard way. It could tolerate large intra-class variations and appropriately discard the corrupted samples. }
\label{fig:concept}
\end{figure*}

To cope with the above problem, a number of tracking approaches~\cite{SVMT, BolmeBDL10, KalalMM10, HenriquesC0B15, SRDCF, CCOT, lukevzivc2017deformable} typically seek to build a variety of robust tracking-by-detection schemes that incrementally learn and update a discriminative appearance model for the target object during tracking within the following optimization framework:
\begin{equation}
\label{CFF}
\min_{\theta}\sum_{k=1}^{t}v_{k}L(g(x_{k},\theta),y_{k}) + \alpha R(\theta).
\end{equation}
Here, $\{(x_{k},y_{k},v_{k})\}_{k=1}^{t}$ denote the sample-label-weight training triplets (collected across consecutive frames) until the current frame $t$, $x_{k}$ and $y_{k}$ contain the set of samples extracted from frame $k$, $v_{k}$ is the sample's weight, $g(x_{k},\theta)$ stands for the discriminative model parameterized by $\theta$ with the regularization term $R(\theta)$ (scaled by $\alpha$), and $L(g(x_{k},\theta),y_{k})$ denotes the loss function. Therefore, such a tracking pipeline is essentially a self-learning process, which makes use of the tracking results from previous frames to determine the latter object localization at the current frame. In this way, the quality of the training samples directly affects the learned appearance model. Since the predictions are just labeled by the tracker itself, the collected training samples are easily corrupted when the predictions are not accurate or when the object is occluded. The reliability of training samples collected online is difficult to guarantee. Therefore, directly learning from the predictions will lead to small tracking errors accumulating and gradually cause model drift. It is a key issue that needs to be tackled to effectively construct and manage a training set that guarantees the success of a tracking model. To address this problem, Danelljan et al.~\cite{SRDCFdecon} propose an adaptive decontamination approach to jointly optimize model parameters and the sample weights as follows:
\begin{equation}
\label{CFF_weight}
\min_{\theta,v\in[0,1]^{t}}\sum_{k=1}^{t}v_{k}L(g(x_{k},\theta),y_{k}) + \alpha R(\theta) + \frac{1}{\mu} \sum_{k=1}^{t} \frac{v_{k}^{2}}{\rho_{k}}.
\end{equation}
Here, $\rho_{k}$ denotes the prior sample weights predefined and $\mu$ is constant. Clearly, the optimal training sample weights in the above formulation are highly correlated with their corresponding sample-specific losses. Namely, the larger loss values generally lead to the smaller sample weights, while the samples with the smaller loss values are greatly encouraged. In essence, such a learning strategy only has one stage, and is more suitable for the learning scenarios with the small intra-class variations of samples and the high inter-class separability. However, in the presence of complicated factors (caused by occlusions, shape deformations, out-of-plane rotations, etc.), the drastic object appearance changes take place, resulting in multi-mode intra-class distributions. In this case, determining the sample weights in a one-pass joint optimization manner is incapable of well capturing the intrinsic multi-mode
diversity properties of the object samples.
Moreover, inspired by the self-paced curriculum learning strategy~\cite{SPL, SPCL},
Supancic and Ramanan~\cite{SPLLTT} sequentially re-learn a SVM model from
previously-tracked ``good'' frames selected according to the SVM objective function
values. As a result, they treat discriminative learning and sample selection separately
without jointly modeling their intrinsic interactions in a unified learning scheme,
resulting in the inflexibility in adapting to complicated time-varying tracking scenarios.

Inspired by the learning process of humans and animals, the self-paced learning is more concerned with the intra-class variations by dynamically selecting easy samples with different patterns~\cite{SPBLC}. It can perform robustly in the presence of extreme outliers or heavy noises~\cite{MengZJ17}. Motivated by the above observations, we in this work seek to build a joint learning agent with the self-paced multi-stage policy of sample selection for robust discriminative learning against relatively large intra-class variations while maintaining inter-class separability. Specifically, we propose a novel time-weighted and detection-guided
self-paced tracking approach, which takes a progressive multi-stage optimization strategy for discriminative learning with the policy of easy-to-hard sample selection in a
time-weighted and detection-guided fashion. At each stage $n$, we iteratively optimize the following objective function in an alternating manner:
\begin{equation}
\label{eq:SP4CF}
\min_{\theta_n,v^{n}\in[0,1]^{t}}\sum_{k=1}^{t}v_{k}^{n}L(g(x_{k},\theta_n),y_{k}) + \alpha R(\theta_n) + f(v^n;\lambda_n).
\end{equation}
Here, $f(v^{n};\lambda_n)$ is a stage-specific sample selection regularizer that controls the learning pace of the sample weights $v_n$. The larger $\lambda_n$ is, the more hard samples with larger loss values are encouraged to be involved in the discriminative learning process. Without loss of generality, suppose we have $N$ learning stages corresponding to $\{(\lambda_1,\theta_1,v^1), \ldots, (\lambda_N,\theta_N,v^N)\}$ such that $\lambda_1<\cdots<\lambda_N$. The iterative optimization procedure in stage $n+1$ takes the learning results $\theta_n$ of stage $n$ as input for the initialization of $\theta_{n+1}$, and is performed until convergence to obtain the optimal $(\theta_{n+1},v^{n+1})$. The above optimization procedure is sequentially repeated until stage $N$. In this way, we have an easy-to-hard evolution process that progressively expands the reliable sample volume involved in discriminative learning, as shown in Figure~\ref{fig:concept}. It starts to train the model from most reliable samples which undergo less appearance variations. As more samples are involved in the learning process, the hard samples (eg. samples with different rotation or motion blur) at the beginning may become much easier and be separated from outliers (eg. background or occluded samples). Such a learning strategy is capable of capturing the multi-grained local distribution structure information of object samples and progressively exploring the intra-class boundaries against the inter-class separability. Meanwhile, it has a better error-tolerating capability because of its progressive learning way. Moreover, we present two variants of the self-paced controller $f(v^n;\lambda_n)$
that take into account more prior information (e.g., temporal importance and detection response confidence).

In summary, the contributions of this work are summarized as follows:
\begin{itemize}
	\item We propose a novel joint discriminative tracking approach with progressive multi-stage optimization. The proposed tracking approach carries out the discriminative learning process in a self-paced learning fashion, which progressively performs easy-to-hard sample selection within a joint optimization framework.
	
	\item We present two novel variants of the self-paced controller, which can respectively incorporate the temporal importance and detection response confidence information into the self-paced learning process. Specifically, they encourage more recent samples or the samples with a good quality of detection response maps to participate in the discriminative tracking process.

	 \item Through progressive learning, our model is able to encode the multi-grained time-varying distribution structure information of object samples during tracking and adapt to intra-class variations together with better error-tolerating power.
\end{itemize}

\section{Related Work}
Visual tracking has been studied extensively over the years. In this paper, we mainly focus on how to make adaptive online sample selection for object tracking. In this section, we first review the relevant works of tracking-by-detection discriminative methods and then give a brief introduction about the self-paced learning method.

Tracking-by-detection approaches have demonstrated the competitive performance in recent years. Usually, they formulate object tracking as a discriminative learning problem to learn a discriminative classifier or regressor in an online manner. Multiple discriminative trackers have been proposed such as multiple instance learning (MIL)~\cite{MILT}, ensemble tracking~\cite{EST}, support vector tracking~\cite{SVMT}, discriminative correlation filter (DCF) based tracking~\cite{BolmeBDL10, HenriquesC0B15, DSST}, and deep learning based tracking~\cite{HCF, SiamT, MDNet, wang2015visual}.

\subsection{DCF based tracking} Discriminative correlation filters (DCF) have been studied as a robust and efficient approach to the problem of visual tracking since Bolme et al.~\cite{BolmeBDL10} initially proposed the Minimum Output Sum of Squared Error (MOSSE) tracker. The DCF is a supervised technique for learning to discriminate the target from the background by solving a linear ridge regression problem. The dense sampling strategy it uses allows the ridge regression problem to be solved extremely efficiently in Fourier domain. Recently, several extensions for DCF based tracking have been proposed to address their inherent limitations to further improve the tracking accuracy. Henriques et al.~\cite{HenriquesC0B15} propose to learn a kernelized version of correlation filter to boost the tracking performance. The methods in~\cite{HenriquesC0B15, GaloogahiSL13} show a notable improvement by learning multi-channel filters on multi-dimensional features such as HOG. For robust scale estimation, Danelljan et al.~\cite{DSST} propose a novel scale adaptive method by learning separate filters for translation and scale estimation. The work ~\cite{eccv/BibiMG16} presents a new formulation by modifying the conventional DCF model to adapt the target response.
Some works are proposed to improve the discrimination of correlation filters by adding color name features~\cite{AdaptColor} or color histograms~\cite{Staple}. Other works, such as BACF~\cite{BACF}, CACF~\cite{CACF}, CFLB~\cite{CFLB}, and SRDCF~\cite{SRDCF} try to address the problem induced by the dense sampling. Specifically, for the mitigation of boundary effects, the methods in BACF and CFLB exploit the background information around the target to add more negative training samples while the approach in SRDCF overcomes this issue by introducing a spatial regularization component within the DCF formulation to penalize filters coefficients depending on their spatial location. More recent works, such as C-COT~\cite{CCOT} and ECO~\cite{ECO}, address the single-resolution feature restriction in the conventional DCF formulation by introducing a new formulation for training continuous convolution filters, which enables efficient integration of multi-resolution features.

\subsection{Deep learning based tracking} Deep learning has recently attracted much attention in machine learning. It has been successfully applied in various computer vision applications including visual tracking. Recent works have either exploited CNN deep features within DCF framework~\cite{CCOT, ECO, DeepSRDCF, HCF, HDT} or builded pure deep architectures~\cite{SiamT, SiamIT, MDNet, CFNet, CREST, PTAV, TSN, DSiam, ACFN, zhang2017visual, wang2015visual} for robust visual tracking. The DCF trackers using deep features could significantly achieve high performance and robustness because of the high discriminative power of such features. For instance, HCF~\cite{HCF} exploits features extracted from pre-trained deep models trained on large datasets and learns multiple CNN layer-wise DCFs to boost the tracking accuracy. MDNet~\cite{MDNet} trains a multi-domain network to learn the shared representation of targets and achieves a great performance improvement. PTAV~\cite{PTAV} presents a novel parallel tracking and verifying framework which coordinates a high accuracy deep tracker and a fast DCF tracker by taking advantage of multi-thread techniques.
In addition, some works~\cite{SiamT, SiamIT} use the siamese network to build template matching based trackers without online updating, which can achieve a good tracking accuracy and simultaneously attain a high tracking speed. Another trend~\cite{CREST, CFNet} in deep trackers is to reformulate DCFs as a differentiable one-layer convolution neural network for end-to-end training, enabling to learn deep features that are tightly coupled to the DCFs. Besides, recurrent neural network and reinforcement learning are recently exploited for tracking. The work~\cite{RTAT} presents a recurrent neural network to capture long-range contextual cues to identify and exploit reliable parts for tracking. The works in~\cite{ADNet, EAST, p-tracker} formulate object tracking as a decision-making process and learn an optimal tracking policy.

\subsection{Model adaptation} As an essential component of online tracking, model update is used to adapt the target appearance changes and should be taken care to avoid model drift. Constructing and managing training samples with high quality and diversity is critical to guarantee the success of online model adaptation. However, there exist some trackers~\cite{DBLP:conf/bmvc/DanelljanHKF14, HenriquesC0B15, Staple} that directly employ model update without considering whether the new collected sample is corrupted or not, which frequently leads to tracking failures. Moreover, most existing trackers employ an explicit training samples management strategy which is independent of the learned model~\cite{BolmeBDL10, KalalMM10, LCT}. Bolme et al.~\cite{BolmeBDL10} propose a  measurement called the Peak to Sidelobe Ratio (PSR) for failure detection and then discard the training sample when the PSR does not meet a certain criterion. Similarly, the greatest peak value in response map is also treated to empirically measure the quality of the detection~\cite{LCT, LMCF}. TLD ~\cite{KalalMM10} uses additional supervision to manage the generation of positive and negative samples to avoid model drift. The Multi-Store Tracker~\cite{MUSTer} maintains a long-term memory of SIFT keypoints for the object and background. SRDCFdecon~\cite{SRDCFdecon} presents a unified formulation for jointly learning the appearance model and the training sample weights, which estimates the quality of training samples based on sample loss. ECO presents a probabilistic generative model (Gaussian Mixture Model) of the training set that obtains a compact description of the samples, which drastically eliminates samples redundancy and reduces memory requirements. The work~\cite{ACFN} designs a deep attentional network to adaptively select the best fitting subset of correlation filters according to the dynamic properties of the tracking target, where each filter covers a specific appearance or dynamic change of the target.

\subsection{Self-paced Learning} Recently, curriculum learning and self-paced learning have gained increasing attention in the field of machine learning. Curriculum learning is a learning paradigm inspired by the intrinsic learning principle of humans/animals~\cite{CL}. Its core idea is to incrementally involve samples into learning, which generally begins with learning easy samples, and then gradually takes more complex samples into consideration. Instead of using the heuristic strategies, Kumar et al.~\cite{SPL} reformulate the key principle of curriculum learning as a concise self-paced learning model. The self-paced model adds a extra regularizer imposed on the sample weights and jointly learns the model parameter and the latent sample weight. Multiple variants of it have been proposed, such as self-paced ranking~\cite{SPLR}, self-paced learning with diversity~\cite{SPLD} and self-paced curriculum learning~\cite{SPCL}.

The works~\cite{CFHSSL, CFSPT} which are most related to ours also employ the self-paced learning to select samples in object tracking. However, different from ours, they employ the one-stage optimization policy and only use loss metric. Overall, our approach differs from the aforementioned method in the aspects of using learning strategies. The aforementioned methods either employ predefined or heuristic strategy or take a one-pass strategy without progressive learning, while our approach is based on a joint optimization scheme that progressively performs sample selection in an easy-to-hard way, which can adapt to intra-class variations together with better error-tolerating power. Moreover, we incorporate detection results as feedback to guide the learning process, which is a complementary measure with respect to the training loss.

\begin{figure*}[!htbp]
\centering
\subfloat[Sample weights]{\includegraphics[width=0.49\textwidth]{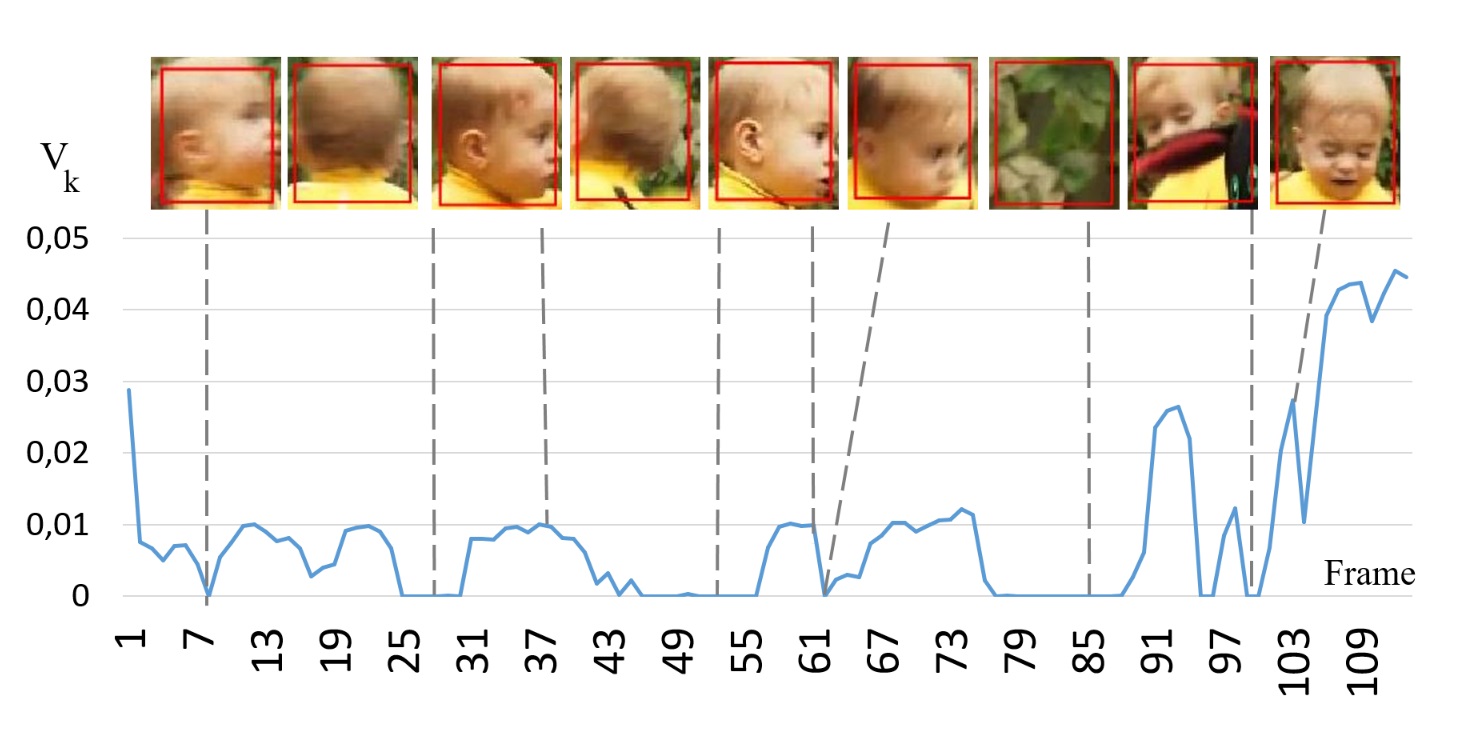}
\label{main}}
\hfil
\subfloat[progressive learning for easy-to-hard sample selection]{\includegraphics[width=0.49\textwidth]{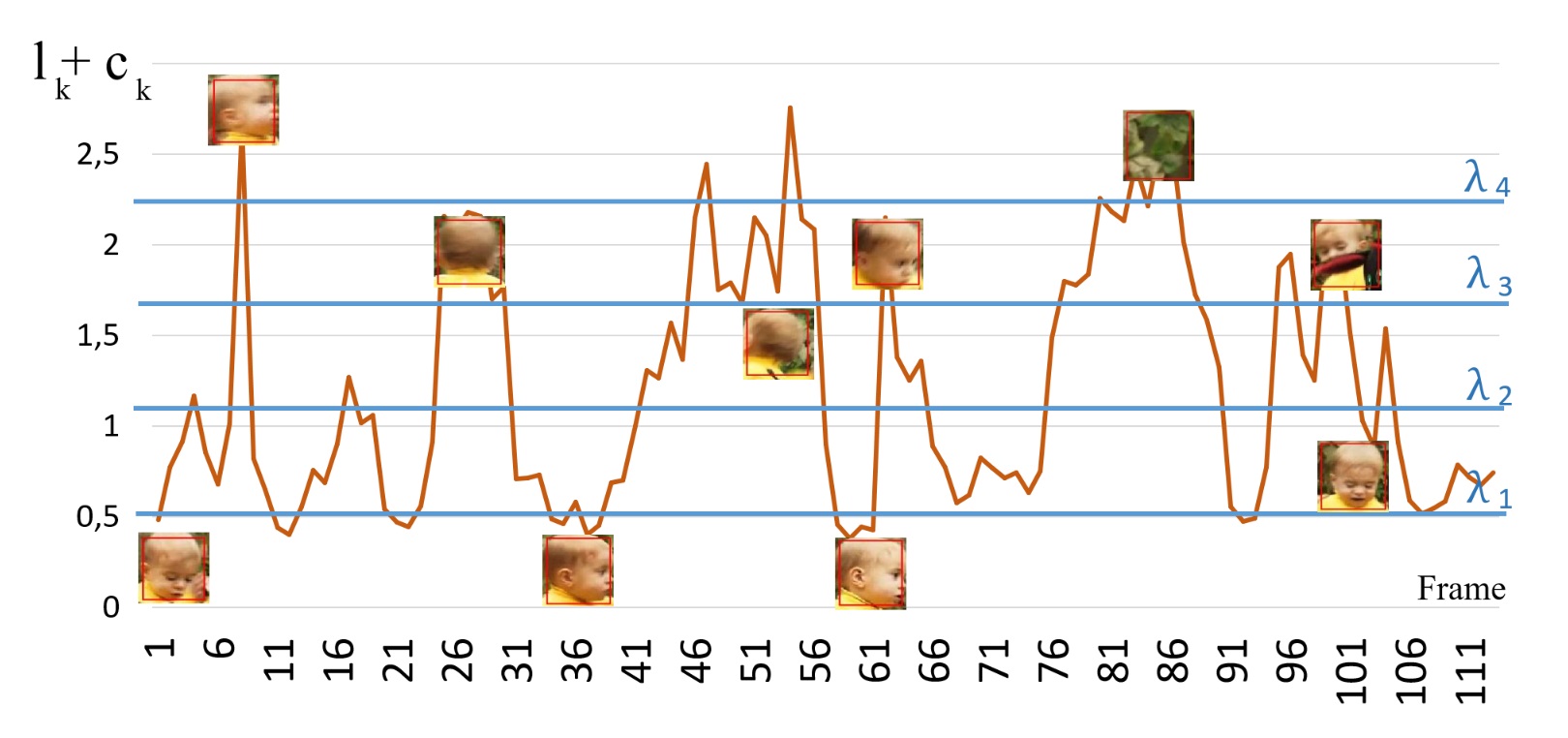}
\label{main2}}
\caption{In part (a) we illustrate a visualization of the $v_k$ values for different samples taken from a video. We can see that the good samples have a high $v_k$ while the ones that are blurred, contain occlusion or are not part of the tracked object have a $v_k$ of 0 eliminating them from being used during the training. In part (b) we plot $l_k + c_k$ of a video for each frame. If we assume a fair trade-off between $l_k$ and $c_k$, then this graph illustrates the threshold at which a sample is eliminated. Since $v_k = 0$ if $c_k + l_k > \lambda$. By tracing different values of $\lambda$ as horizontal lines, we can see which frames are considered during which training stage. we can see from the figure that the samples in the lower part of the graph (those with that are considered first in the training) are easier than those in the high part of the graph which contains blurred, occluded and corrupted samples.}
\label{fig:overview3}
\end{figure*}

\section{Our Approach}
\subsection{Overview}
Discriminative visual tracking is an online learning problem, which trains a discriminative appearance model as shown in Eq.~\eqref{CFF}, given the training samples $\{(x_{k},y_{k})\}_{k=1}^{t}$ collected from the video sequence.  In the presence of complicated factors (caused by partial occlusion, shape deformation, out-of-plane rotation, etc.), the target appearance undergoes drastic changes, resulting in multi-mode intra-class distributions. The training samples collected from the predictions are just labeled by the tracker itself and are difficult to guarantee their quality. Directly learning form the samples will often accumulate small tracking errors and finally cause model drift. How to select samples to train the appearance model against relatively large intra-class variations while maintaining inter-class separability is a key issue to be tackled for tracking. In addition, visual tracking is a sequential decision making task, samples from different frames should have different impacts on detection in the current frame. In the learning process, the temporal importance information should be considered for sample selection. Especially, the impact of samples in recent frames should be emphasized and assigned more weights in the subsequent predictions. Another factor needing to be considered is the measure of the sample quality. Apart from the training loss used for evaluating the sample quality, the detection response map may also be a cue to reveal the sample quality. The prediction is typically considered to be reliable and accurate, if the detection map has only one sharp peak with low variance. Therefore, the detection confidence map can be adopted as another measure to estimate the sample quality. In conclusion, the design of sample selection for tracking should take into account both the temporal importance information and the detection confidence.

Therefore, we in this work present a self-paced multi-stage learning formulation for discriminative learning against relatively large intra-class variations while maintaining inter-class separability, as formulated in Eq.~\eqref{eq:SP4CF}. In the following, we will first introduce the self-paced learning with the standard self-paced controller and then present two novel variants of the self-paced controller, which can respectively incorporate the temporal importance and detection response confidence information into the self-paced learning process. For brevity and conciseness, we use $l_{k}$ to denote the sample loss $L(f(x_{k},\theta),y_{k})$ in the following section.

The function $f(v^{n};\lambda_n)$ in Eq.~\eqref{eq:SP4CF} is a self-paced regularizer that controls the stage-specific sample selection. Typically, a self-paced regularizer with a linear weighting scheme in conventional self-paced learning~\cite{SPCL, SPL} is defined as:
\begin{equation}\label{eq:LWScheme}
f(v^{n};\lambda_n) = \lambda_n \sum_{k=1}^{t}(\frac{1}{2}(v_{k}^{n})^{2} - v_{k}^{n}).
\end{equation}
\noindent Here, $\lambda_n$ denotes the learning pace, which gradually increases and involves more samples into model training with the learning process going. When substituting Eq.~\eqref{eq:LWScheme} into Eq.~\eqref{eq:SP4CF} and fixing the appearance parameters $\theta_n$, Eq.~\eqref{eq:SP4CF} becomes a convex function of $v^{n}$ and thus the closed-form optimal solution for the sample weights $v^{n}$ can be derived as:
\begin{equation}\label{eq:LWSolution}
v_{k}^{n} = \left\{
\begin{array}{lcl}
1 - \frac{l_{k}}{\lambda_n}, && l_{k} < \lambda_n; \\
0, && l_{k} \geq \lambda_n.
\end{array}\right.
\end{equation}
As shown in Eq.~\eqref{eq:LWSolution}, every sample is treated to be equally important, assigning their weights only by their losses. Figure~\ref{main} illustrates which samples are considered and which are ignored based on the values of $v_k$. It can be seen that the samples with blur or occlusion have $v_k = 0$ causing them to be ignored by the training process while those that are easy to learn have higher weights.

\subsection{Temporal importance integration}
In visual tracking, the target can undergo unpredictable transformations over time. To adapt to drastic appearance changes, recent samples should be assigned more weights while the older ones should get forgotten gradually. Instead of treating every sample equally in the original self-paced regularizer in Eq.~\eqref{eq:LWScheme}, we encode the temporal importance information for tracking into the self-paced regularizer $f(v^{n};\lambda_n)$  to guide the learning process. The time-weighted self-paced regularizer $f(v^{n};\lambda_n)$ is formulated as:
\begin{equation}\label{eq:TPR}
  f(v_n;\lambda_n) = \lambda_n\sum_{k=1}^{t}(\frac{1}{2}\frac{{(v_{k}^{n})}^{2}}{\rho_{k}} - v_{k}^{n}).
\end{equation}
Here, $\rho_{k}$ denotes the temporal importance of the sample $x_{k}$. It has higher values for the more recent samples while the weights of previous samples are decayed exponentially over time. By substituting Eq.~\eqref{eq:TPR} into Eq.~\eqref{eq:SP4CF}, we can obtain the following objective function for the learning stage $n$:
\begin{equation}\label{eq:SPTPR}
  \min_{\theta_n,v^{n}\in[0,1]^{t}}\sum_{k=1}^{t}v_{k}^{n}l_{k} + \alpha R(\theta_n) + \lambda_n\sum_{k=1}^{t}(\frac{1}{2}\frac{(v_{k}^n)^{2}}{\rho_{k}} - v_{k}^{n}).
\end{equation}
When fixing the appearance parameters $\theta_{n}$, Eq.~\eqref{eq:SPTPR} is a convex function with respect to $v^{n}$ and thus the close-formed optimal solution for the sample weights $v^{n}$ can be derived as:
\begin{equation}\label{eq:TPRS}
  v_{k}^{n} = \left\{
  \begin{array}{lcl}
    \rho_{k} - \frac{l_{k}}{\lambda_n}\rho_{k}, && l_{k} < \lambda_n; \\
    0, && l_{k} \geq \lambda_n.
  \end{array}\right.
\end{equation}
As seen in Eq.~\eqref{eq:TPRS}, higher values of $\rho_{k}$ will drive $v_{k}^{n}$ up. Therefore, the recent and reliable samples can be emphasized in the learning process by assigning higher temporal confidence $\rho_{k}$. In our work, we define $\rho_{k}$ as:
\begin{equation}\label{eq:PK}
  \rho_{k} = (1-\eta)\rho_{k+1}.
\end{equation}
where $\eta$ is the learning rate. This drives the learning agent to put more weight on recent frames and let the effect of pervious frames decay exponentially over time.

\subsection{Detection confidence integration}
In discriminative tracking, the distribution of the response map can be used as a measurement to predict the location accuracy~\cite{BolmeBDL10, LCT, LMCF}. When the detected target is well matched to the correct object, the ideal response map should have only one sharp peak and low values in all other locations. The sharper the response peak is, the better the location accuracy is attained. The presence of distractions in the image during the tracking often leads into multiple peaks instead of one, which may be due to some factors like partial occlusion.

Instead of only using the training loss to estimate the sample quality, we incorporate detection results as feedback to guide the learning process, which is a complementary measure with respect to the training loss. We reformulate Eq.~\eqref{eq:TPR} by integrating with the detection confidence as:
\begin{equation}\label{eq:DFR}
  f(v^{n};\lambda_n) = \lambda_n\sum_{k=1}^{t}(\frac{1}{2}\frac{{(v_{k}^{n})}^{2}}{\rho_{k}} - v_{k}^{n}) + \xi\sum_{k=1}^{t}c_{k}v_{k}^{n}.
\end{equation}
\noindent Here, $c_{k}$ denotes the detection quality in the frame $t_{k}$. The smaller $c_{k}$ is, the more reliable the sample $x_{k}$ is. $\xi \geq 0$ represents the trade-off between the training loss and detection score. Furthermore, we can reformulate the optimization objective Eq.~\eqref{eq:SPTPR} as:
\begin{equation}\label{eq:SPDFR}
  \hspace{-1em}\min_{\theta_n,v^{n}\in[0,1]^{t}}\sum_{k=1}^{t}v_{k}^{n}l_{k} + \alpha R(\theta_n) + \lambda_n\sum_{k=1}^{t}(\frac{1}{2}\frac{(v_{k}^{n})^{2}}{\rho_{k}} - v_{k}^{n}) + \xi\sum_{k=1}^{t}c_{k}v_{k}^{n}.
\end{equation}
When fixing the model parameters $\theta_n$, we can derive the closed-form optimal solution for the sample weights as:
\begin{equation}\label{eq:DFRS}
  v_{k}^{n} = \left\{
  \begin{array}{lcl}
    \rho_{k} - \frac{l_{k} + \xi c_{k}}{\lambda_n}\rho_{k}, && l_{k} + \xi c_{k}< \lambda_n; \\
    0, && l_{k} + \xi c_{k} \geq \lambda_n.
  \end{array}\right.
\end{equation}
\noindent It can be seen from Eq.~\eqref{eq:DFRS} that the frame weights are determined by both training loss and detection confidence. Smaller values of $c_{k}$  will increase the sample's weight. If we assume that $\xi = 1$, then Figure~\ref{main2} illustrates how the self-paced learning progresses as we increase $\lambda_n$, the samples with low $c_k$ and $l_k$ values are easy and are absorbed early in the training while those that have higher values are eliminated and only added later when $\lambda_n$ becomes high enough.

Furthermore, we propose two criteria to estimate the detection confidence $c_{k}$. The first one is the maximum score $g_{max1}$ of the response map. The second one is a novel criterion called peak ratio ($pr$) which is formulated as:
\begin{equation}\label{eq:PR}
  pr = \frac{g_{max2}}{g_{max1}}.
\end{equation}
\noindent Where $g_{max1}$ and $g_{max2}$ denote the highest peak's value and the second highest peak's value respectively. Then we define the detection confidence $c_{k}$ as:
\begin{equation}\label{eq:CK}
  c_{k} = \left\{
  \begin{array}{lcl}
    0, && \text{if}\;pr \leq \beta_{1} \land g_{max1} > \beta_{2};\\
    pr, && \text{otherwise}.
  \end{array}\right.
\end{equation}
\noindent It can be seen from Eq.~\eqref{eq:CK} that $c_{k}$ will be small when the response map has only one sharp peak, which means the prediction is of accuracy and more sample weights will be assigned to $v_{k}^{n}$.

\begin{algorithm}
	\caption{The tracking algorithm with our progressive learning strategy}
	\label{alg:outline}
	\begin{algorithmic}[1]
		\REQUIRE
		Frames $\{I_{t}\}_{t=1}^{T}$, initial location $p_{1}$, initial learning pace $\lambda$, scale ratio $\mu$,  learning stages $N$.
		\ENSURE
		Target locations of each frame $\{p_{t}\}_{t=2}^{T}$;
		\FOR{$t = 1 \to T$}
		\STATE Utilize the tracker to estimate the target position $p_{t}$
        \STATE Extract sample $(x_{t}, y_{t})$ to training set $\{(x_{t}, y_{t})\}_{1}^{t-1}$.
		\STATE Estimate detection confidence $c_{t}$ with Eq.~\eqref{eq:CK}.
        \STATE Update prior sample weights $\{\rho_{k}\}_{k=1}^{t}$ as Eq.~\eqref{eq:PK}.
		\FOR{$n = 1 \to N $}
		\WHILE{not converged}
		\STATE \textbf{update} $v^{n}$\textbf{:} Fix $\theta_n$, estimate $v^{n}$ with Eq.~\eqref{eq:DFRS}.
		\STATE \textbf{update} $\theta_n$\textbf{:} Fix $v^{n}$, obtain $\theta_n$ by optimizing Eq.~\eqref{CFF}.
        \ENDWHILE		
		\STATE Increase the learning pace $\lambda_n = \mu\lambda_{n-1}, \mu > 1$.
	    \ENDFOR	
		\RETURN position $p_{t}$;
		\ENDFOR
	\end{algorithmic}
\end{algorithm}

\subsection{Optimization}
Based on the self-paced learning strategy, our approach takes a progressive multi-stage learning strategy for discriminative learning with the policy of easy-to-hard sample selection. At each stage $n$, we optimize Eq.~\eqref{eq:SPDFR} to obtain the stage-specific appearance model parameters $\theta_{n}$ and the sample weights $v^{n}$. It can be verified that the Eq.~\eqref{eq:SPDFR} is biconvex when the loss function is convex, and most discriminative tracking methods like DCF or SVM use a convex loss function $l_{k}$. To solve the biconvex problems, the Alternative Convex Search (ACS)~\cite{ACS} method is utilized for efficient optimization. As a result, we can divide the biconvex optimization~\eqref{eq:SPDFR} into the following two subproblems by optimizing the parameters $\theta_n$ of appearance model and the sample weights $v^{n}$ in an alternating manner:

\noindent \textbf{Update} $v^{n}$ (with $\theta^{n}$ fixed): the optimization problem~\eqref{eq:SPDFR} is reduced as:
\begin{equation}\label{eq:OFW}
\min_{v^n\in[0,1]^{t}}\sum_{k=1}^{t}v_{k}^nl_{k}+\lambda_n \sum_{k=1}^{t}(\frac{1}{2} \frac{{v_{k}^{n}}^2}{\rho_{k}} - v_{k}^n ) + \xi \sum_{k=1}^{t}c_{k}v_{k}^n
\end{equation}
The Eq.~\eqref{eq:OFW} is convex with respect to $v^{n}$. By setting the derivative with respect to $v^{n}$ to zero, the closed-form solution for the sample weights $v^{n}$ can be obtained as in Eq.~\eqref{eq:DFRS}.

\noindent \textbf{Update} $\theta_{n}$ (with $v^{n}$ fixed): the optimization problem in Eq.~\eqref{eq:SPDFR} is reduced as the standard objective function for discriminative tracking in Eq.~\eqref{CFF}, which can be solved by the standard tracking solvers.

\subsection{The tracking framework}
The proposed formulation is flexible and can be integrated into a number of discriminative tracking methods. The tracking algorithm with our progressive learning strategy is summarized in Algorithm \ref{alg:outline}. Specifically, when the frame $t$ arrives, we first utilize the tracker to estimate the object position. Then the new samples $\{x_{t}, y_{t}\}$ are extracted and added to the training set. After that, we update the sample temporal importance $\rho_{k}$ as defined in Eq.~\eqref{eq:PK}. Then we evaluate the detection confidence $c_{k}$ for the new frame $x_{t}$ with Eq.~\eqref{eq:CK}. Lastly, we employ a multi-stage learning mechanism to update the tracking model $\theta$, which absorbs samples into the training process from easy to hard by gradually increasing the learning pace. In each learning stage, we minimize the joint loss in Eq.~\eqref{eq:SPDFR} by alternately solving the model parameters $\theta_n$ and the sample weights $v^{n}$. We employ the same sample replacement strategy as in SRDCFdecon~\cite{SRDCFdecon}, in which if the number of samples exceeds $T$, the sample that has the smallest weight $v_{k}^{n}$ will be removed from the training set.
\begin{figure*}[!htbp]
\centering
\subfloat{\includegraphics[width=0.45\textwidth]{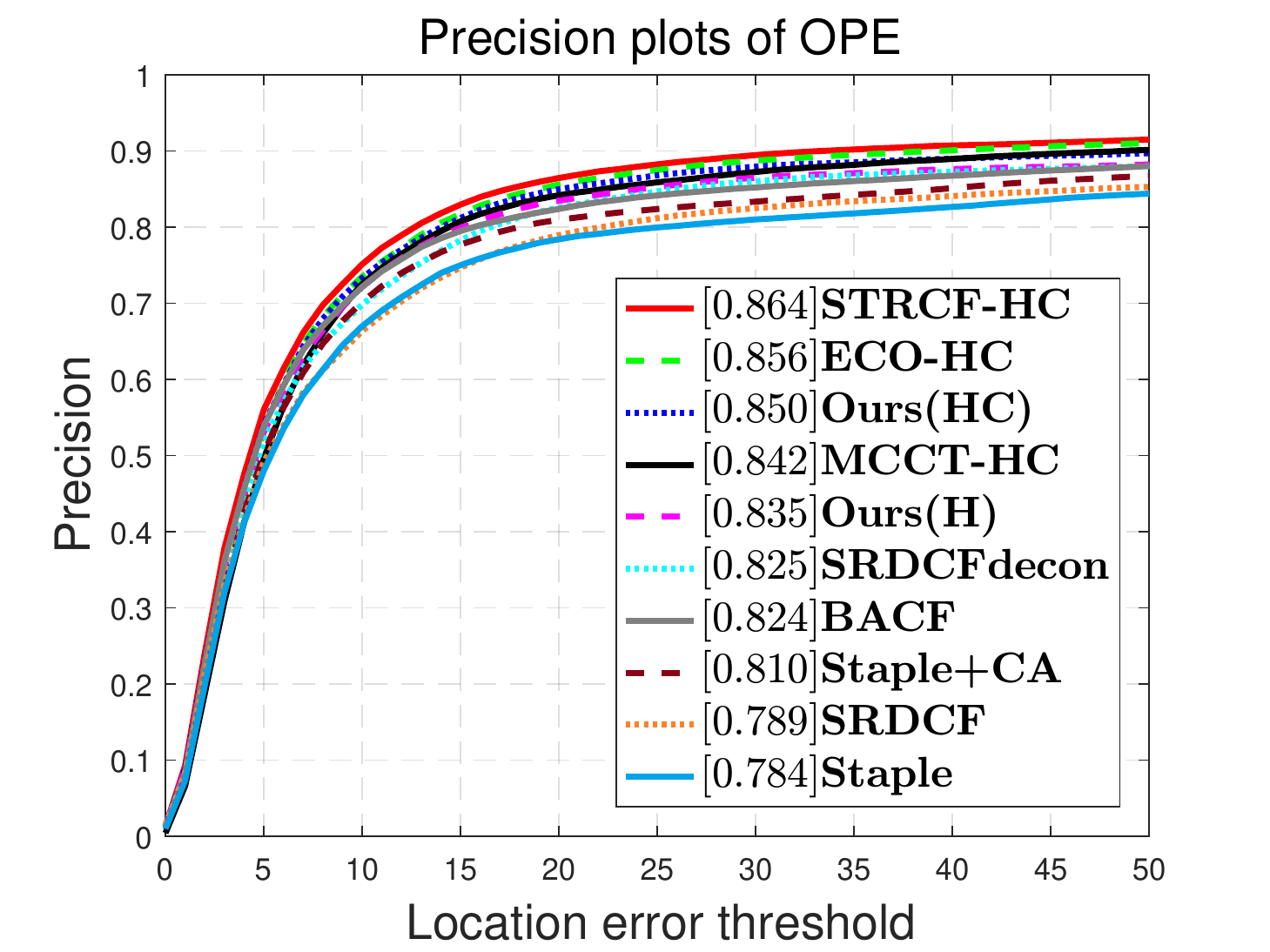}
\label{fig_precision}}
\hfil
\subfloat{\includegraphics[width=0.45\textwidth]{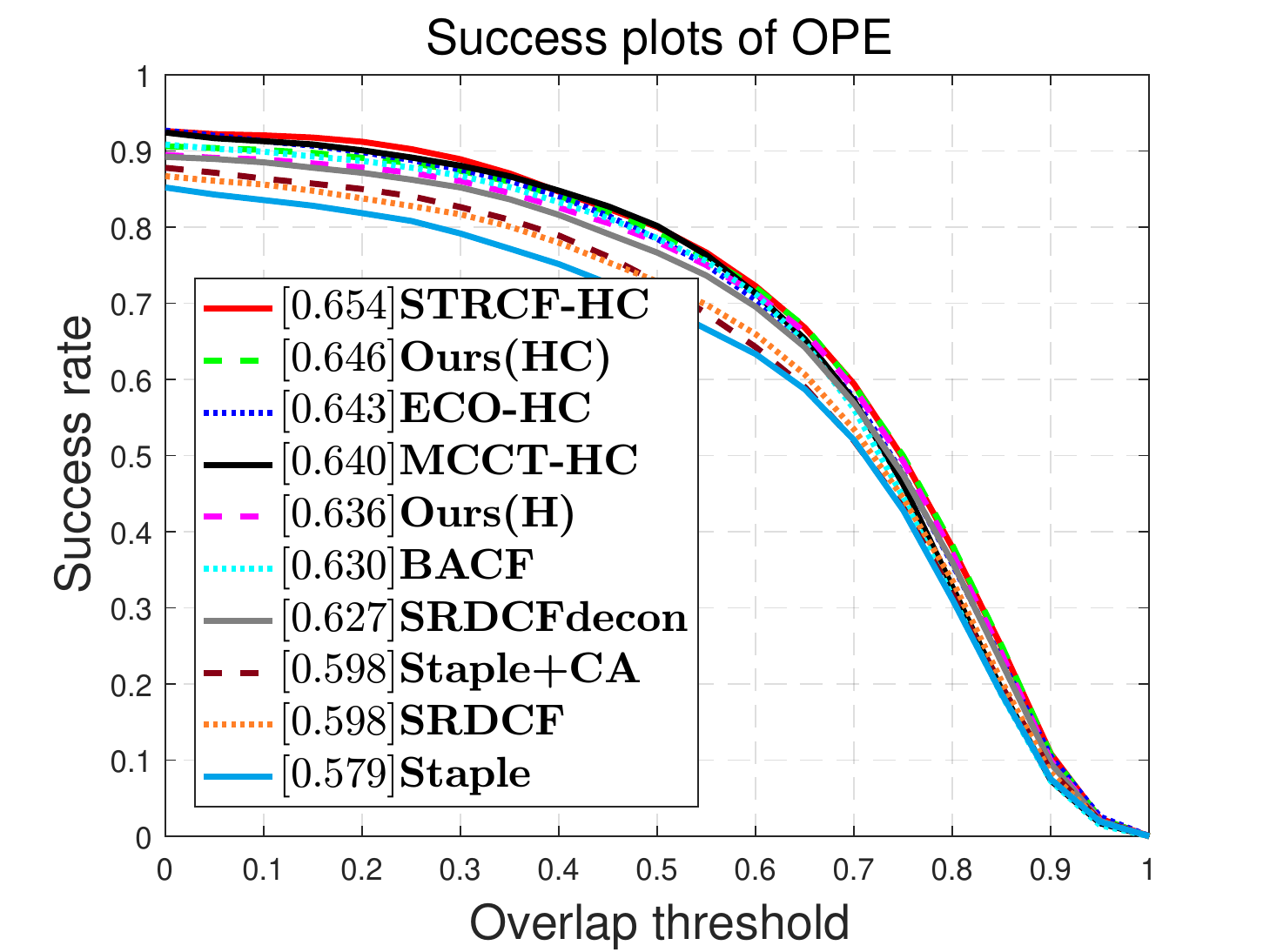}
\label{fig_success}}
\caption{Precision and success plot over all the 100 sequences using one-pass evaluation on the OTB-2015 dataset. Our method can achieve competitive performance with the state-of-art trackers.}
\label{fig:stateofart}
\end{figure*}
\section{Experiments}
To validate the effectiveness of our proposed framework, we integrate it with the popular tracker SRDCF. We then perform extensive experiments on the popular object tracking benchmarks OTB-2015~\cite{OTB2015}, Temple Color~\cite{TempleColor}, VOT-2015~\cite{VOT2015}, VOT-2016~\cite{VOT2016} and VOT-2017~\cite{VOT2017}.

\subsection{Experimental Setup}
\subsubsection{Parameter settings} In our experiments, the settings of parameters depend on the experimental performance and are kept fixed for all the tracking videos in one dataset. Specifically, the prior weights $\rho$, the number of sample buffer $T$, and the features used are set to the same as the trackers SRDCF~\cite{SRDCF} and SRDCFdecon~\cite{SRDCFdecon}. In our optimization, at each stage $n$, our approach optimizes Eq. (11) to obtain the stage-specific model parameters $\theta_{n}$ and the sample weights $v^{n}$. The solution of $v^{n}$ can be directly derived by Eq. (12), which is highly efficient. The complexity of solving the parameters $\theta_{n}$ depends on the baseline training procedure, which dominates the computational cost. To balance the tradeoff between tracking accuracy and computational efficiency, we set the number of the learning stages $N = 3$ and the number of ACS iterations at each learning stage to 1, and conduct model updates every 6 frames. Our tracker is implemented in MATLAB on an Intel 3.5 GHz CPU with 32G RAM and run about 5 frames per second.

\subsubsection{Evaluation metrics}
For the OTB-2015 and Temple Color datasets, the tracking quality is measured by precision and success metric as defined in OTB-2015. The precision metric indicates the rate of frames whose center location are within some certain distances with the ground truth location. The success shows overlap ratio between estimated bounding box and ground truth. We report the average of precision score at 20 pixels threshold (PS) and the average of the area under curve (AUC) of success plot in one-pass evaluation.

\subsection{Internal Analysis of the proposed approach}
\subsubsection{Impacts of the different self-paced controllers}
To demonstrate the effectiveness of our proposed formulation, we evaluate our framework on the OTB-2015 benchmark by incorporating different self-paced controllers. For clarity, we here denote the baseline tracker with a linear weight regularizer in Eq.~\eqref{eq:LWScheme} as SPL, with temporal knowledge integration in Eq.~\eqref{eq:TPR} as SPL-TKI and with temporal and detection confidence knowledge in Eq.~\eqref{eq:DFR} as SPL-DCI.

The tracking results are summarized in Table~\ref{tb:baseline}. According to the experimental results, our proposed method can significantly improve the tracking performance compared with the baseline tracker. Specifically, the basic SPL tracker achieves about 2.7\% and 1.9\% improvement with PS and AUC on the OTB-2015 dataset. The SPL-TKI further improves the performance by a relative gain of 1.2\% and 1.3\% in PS and AUC. Additionally the SPL-DCI achieves a PS score of 0.835 and an AUC score of 0.636, leading to a final relative gain of 5.8\% in PS and 6.4\% in AUC compared to the baseline. Thus, our proposed joint learning framework SPL and two variants of the self-paced controllers SPL-TKI and SPL-DCI can significantly boost the tracking performance.

\begin{table}[!htbp]
\caption{Analysis of our approach with different self-paced regularizers on the OTB-2015.}
\label{tb:baseline}
\centering
\begin{tabular}{|l|c|c|c|c|}
\hline
& Baseline~\cite{SRDCF} & SPL & SPL-TKI & SPL-DCI \\
\hline
Precision  & 0.789 &   0.816  & 0.828 & \textbf{0.835} \\
\hline
Success    & 0.598 &   0.617  & 0.630 & \textbf{0.636} \\
\hline
\end{tabular}
\end{table}

\begin{figure}[!htbp]
\centering
    \includegraphics[width=0.4\textwidth]{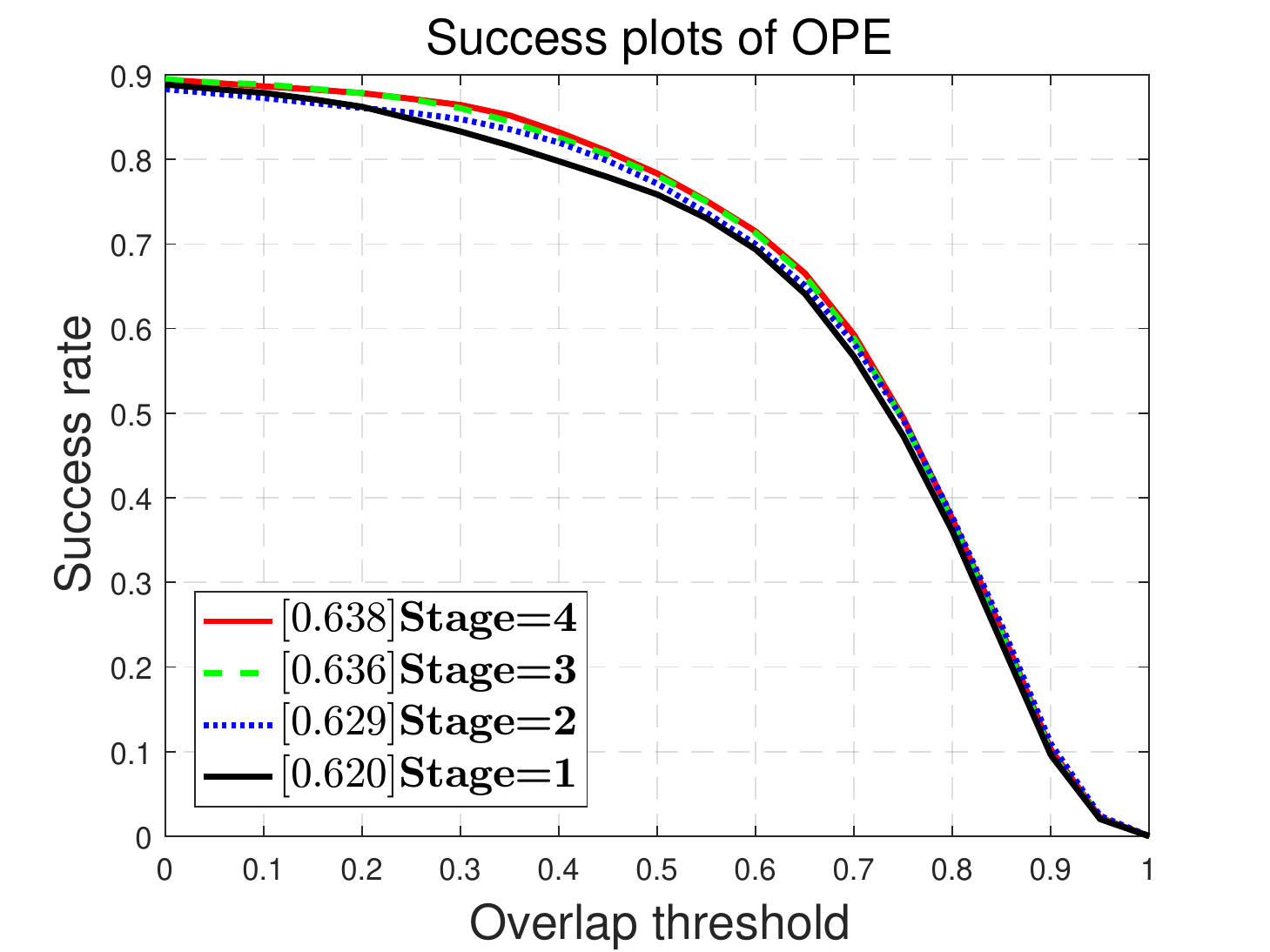}
    \caption{The precision plot using one-pass evaluation on the OTB-2015 dataset in terms of four different settings of the number of learning stages.}
\label{fig:ablation_stages}
\end{figure}

\begin{figure}[!htbp]
\centering
        \includegraphics[width=0.4\textwidth]{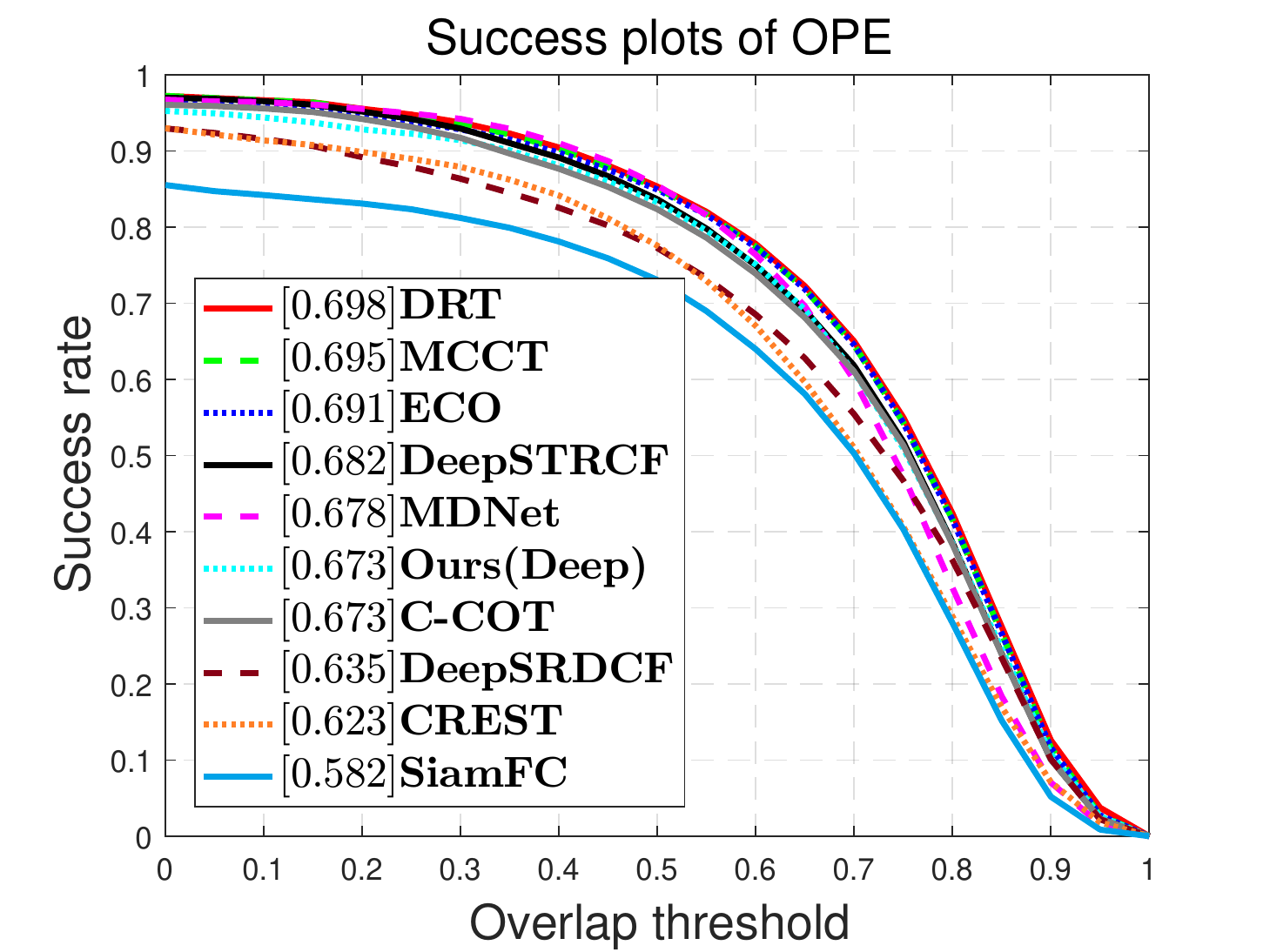}
    \caption{The comparison of the overlap success plots with the deep features-based trackers on OTB-2015 dataset.}
\label{fig:extra_hog}
\end{figure}

\subsubsection{Effect of the learning stages $N$}
We further analyze the effect of the learning stages $N$ on the tracking performance. As seen in Fig.~\ref{fig:ablation_stages}, as the stage number $N$ increases, the performance of the tracker gradually gets better, demonstrating the effectiveness of the multi-stage policy of sample selection. In our experiments, to balance the speed and the performance of the tracker, the stage number $N$ is set to 3.

\begin{table}[h]
\caption{The effect on the tracking accuracy and speed in terms of different model update intervals on OTB-2015 dataset.}
\label{tb:frequency}
\setlength{\tabcolsep}{0.8mm}
\begin{center}
\begin{tabular}{|l|c|c|c|c|c|c|c|c|}
\hline
Intervals & 1 & 2 & 3 & 4 & 5 & 6 & 7 & 8 \\
\hline
AUC score & 0.639 & 0.635 & 0.637 & 0.634 & 0.634 & 0.636 & 0.631 & 0.628 \\
\hline
Frames per second & 2.3 & 3.2 & 3.9 & 4.3 & 5.1 & 5.2 & 5.6 & 6.1 \\
\hline
\end{tabular}
\end{center}
\end{table}

\subsubsection{Effect of the model update intervals}
Table~\ref{tb:frequency} shows our method's accuracy and speed in terms of different model update intervals. As the model update interval increases, the tracking accuracy slightly goes down while the tracking speed gradually goes up. When we perform model update every six frames, our method can achieve a competing performance while maintaining a relative high tracking speed. Therefore, we choose to update the model every six frames.

\begin{figure}[h]
\centering
        \includegraphics[width=0.4\textwidth]{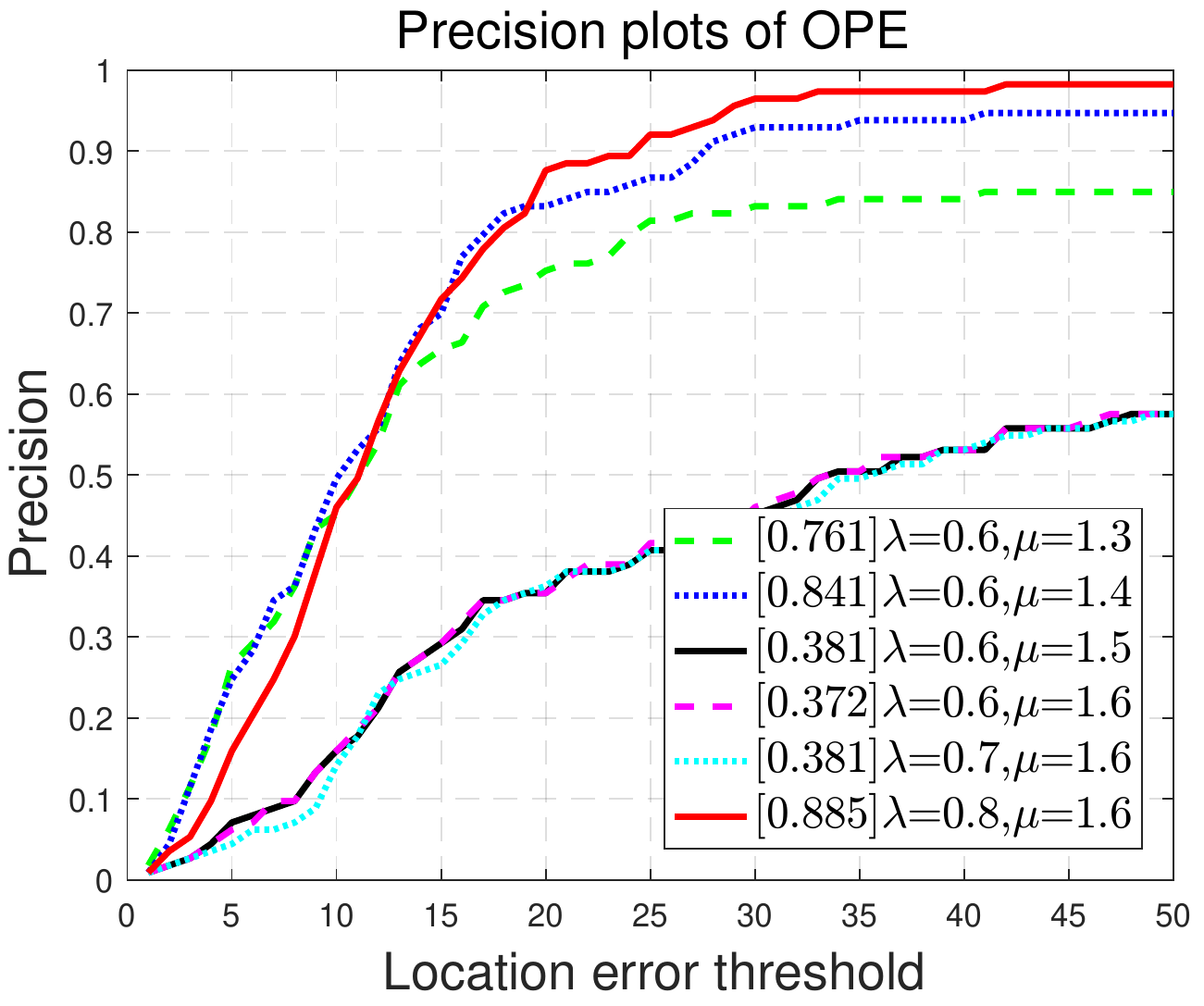}
    \caption{Precision plot over the \textit{dragonbaby} sequence using one-pass evaluation in terms of different settings of the initial learning pace $\lambda$ and scale ratio $\mu$.}
\label{fig:ablation_paces}
\end{figure}

\subsubsection{Effect of the initial learning pace $\lambda$ and scale ratio $\mu$}
Fig.~\ref{fig:ablation_paces} shows the tracking accuracy in terms of different settings of $\lambda$ and $\mu$. It can be seen from Fig.~\ref{fig:ablation_paces} that the choice of $\lambda$ and $\mu$ is of importance to the accuracy of our tracker. Theoretically, the proper settings of the $\lambda$ and $\mu$ should depend on the scale of both the sample loss and confidence score of each video sequence.

\subsection{OTB-2015 Dataset}
In this section, we carry out experiments on the OTB-2015 dataset with comparisions to the astate-of-art tracking methods including MOSSE~\cite{BolmeBDL10}, TLD~\cite{KalalMM10}, TGPR~\cite{TGPR}, CSK~\cite{CSK}, KCF~\cite{HenriquesC0B15}, DSST~\cite{DSST}, SAMF~\cite{SAMF}, MEEM~\cite{MEEM}, Staple~\cite{Staple}, SRDCF~\cite{SRDCF}, SRDCFdecon~\cite{SRDCFdecon}, DeepSRDCF~\cite{DeepSRDCF}, MCCT~\cite{MCCT}, ECO~\cite{ECO}, C-COT~\cite{CCOT}, SiamFC~\cite{SiamT}, CFNet~\cite{CFNet}, CREST~\cite{CREST}, MDNet~\cite{MDNet} and STRCF~\cite{STRCF}.
\subsubsection{Comparison with hand-crafted based trackers}
A comparison with state-of-art trackers on the OTB-2015 is shown in Figure~\ref{fig:stateofart}. For clarity, only top 10 trackers are reported. It can be seen that our tracker using HOG features achieves competitive results with a PS score of 83.5\% and an AUC score of 63.6\%, outperforming the baseline SRDCF by 4.6\% and 3.8\%. It further surpasses SRDCFdecon and BACF. The version using HOG and Color Names (HC) features also outperforms MCCT-HC and ECO-HC on AUC metric.

\begin{figure*}[!htbp]
\centering
\subfloat{\includegraphics[width=0.32\textwidth]{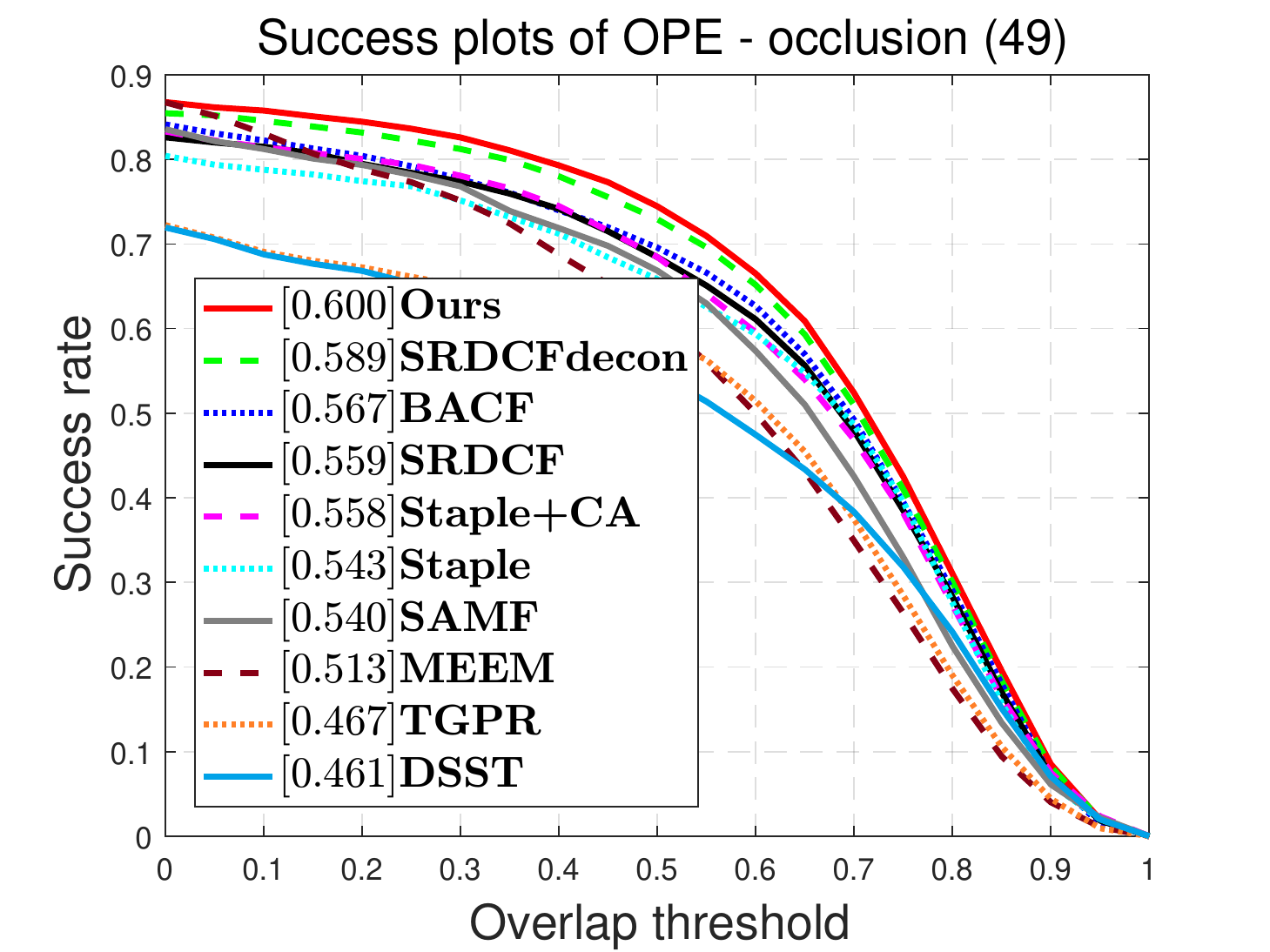}
\label{auc_fast_motion}}
\hfil
\subfloat{\includegraphics[width=0.32\textwidth]{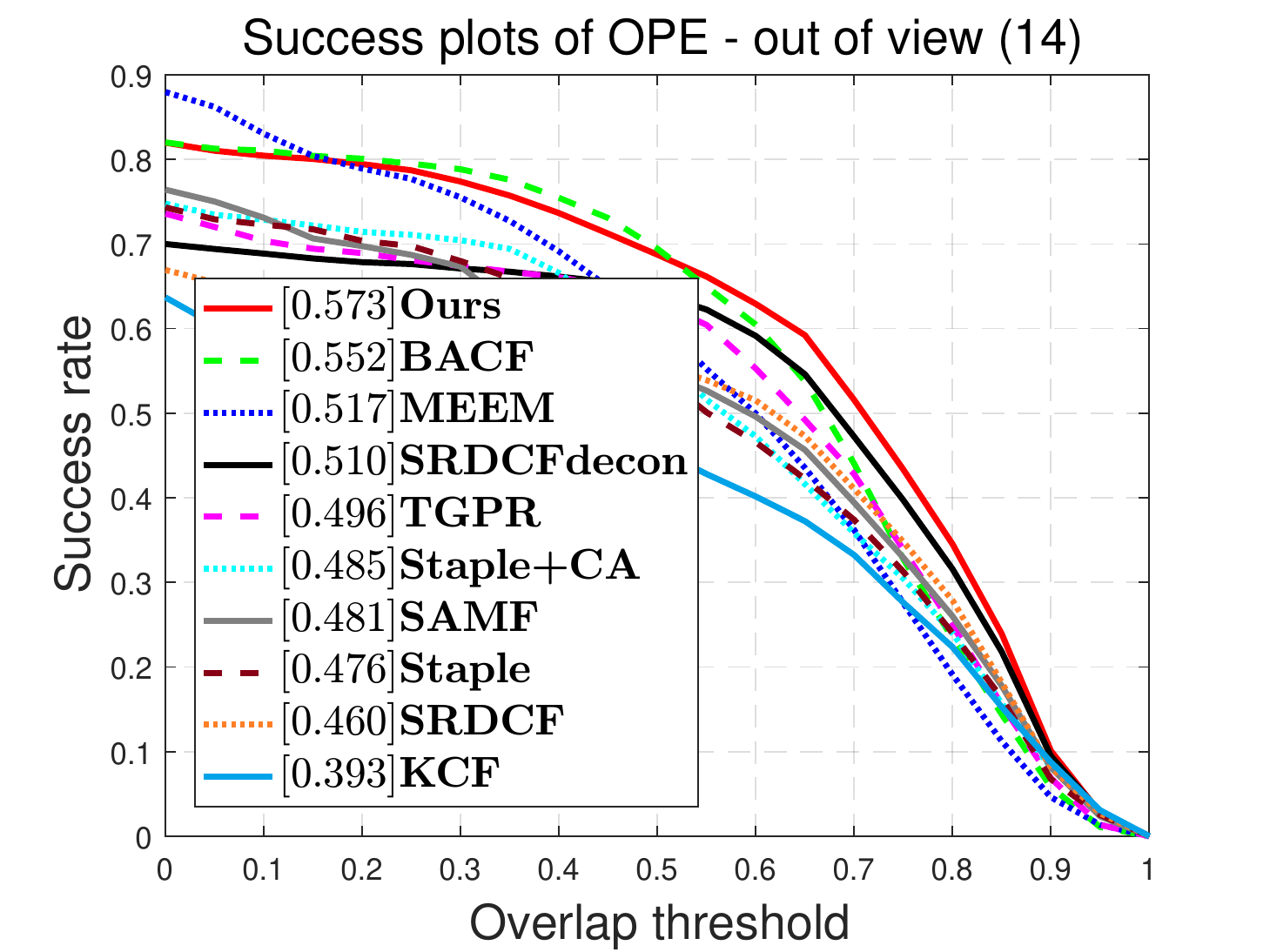}
\label{auc_outofview}}
\hfil
\subfloat{\includegraphics[width=0.32\textwidth]{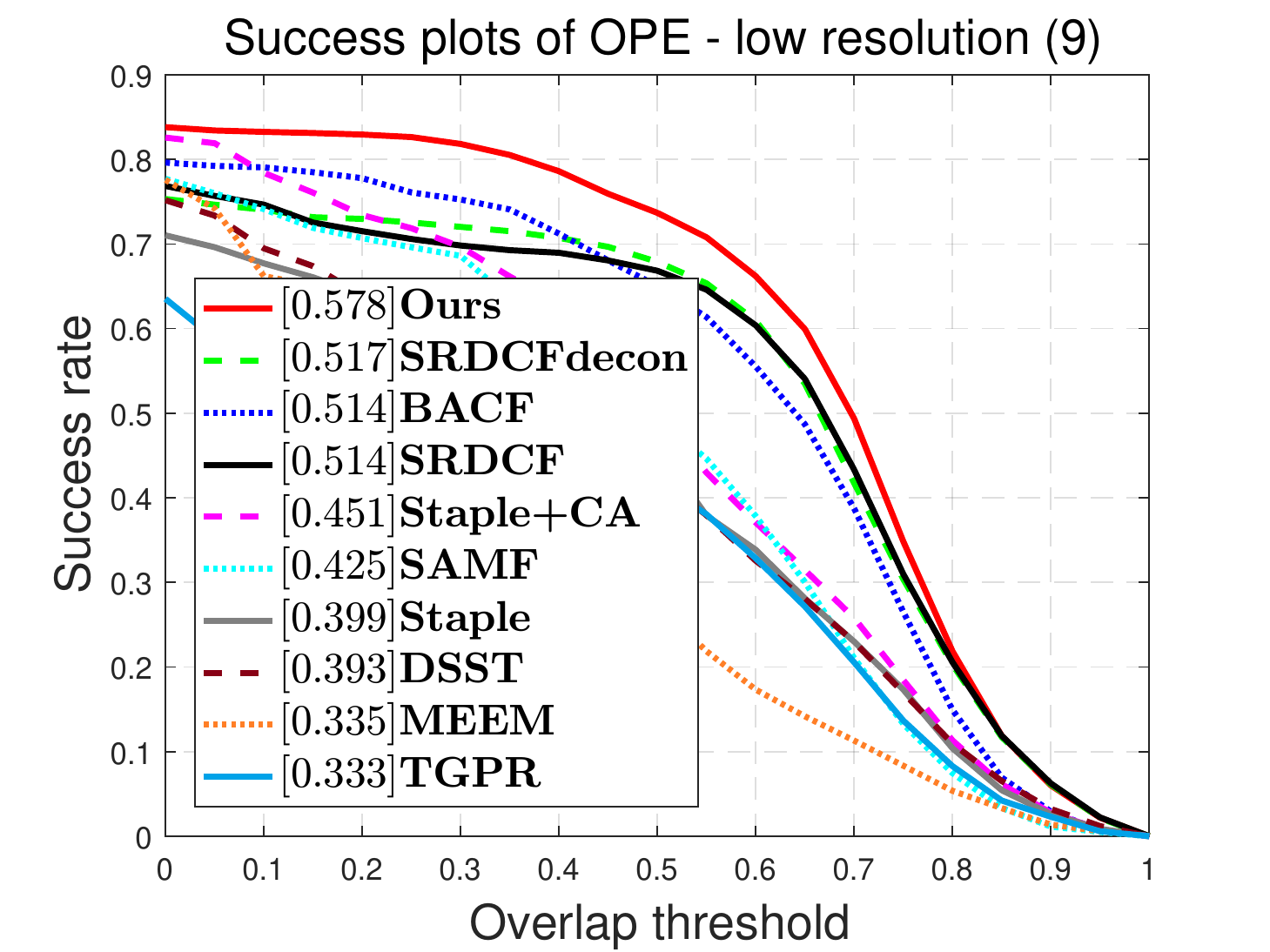}
\label{auc_low_resolution}}
\hfil
~
\centering
\subfloat{\includegraphics[width=0.32\textwidth]{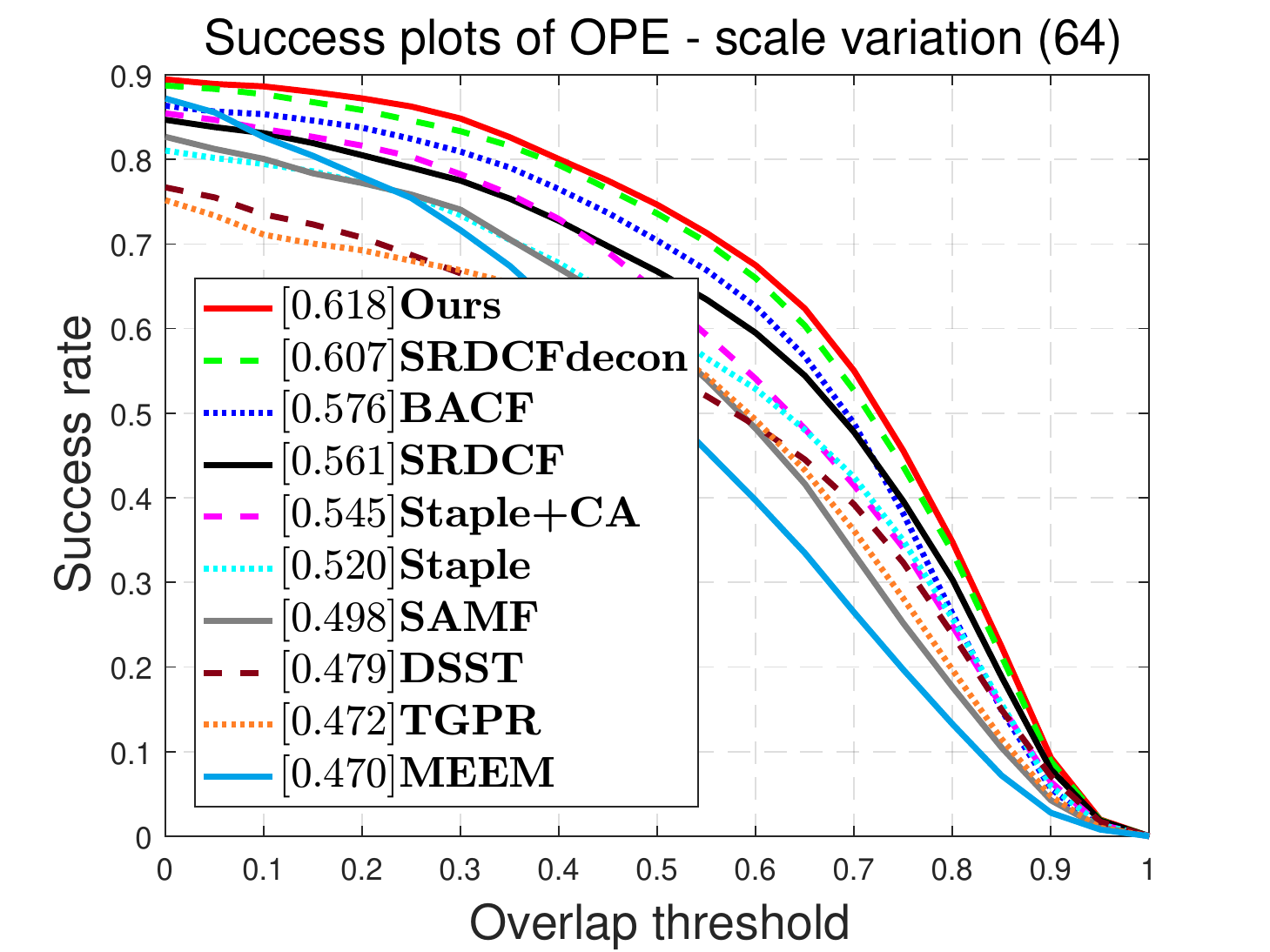}
\label{auc_motion_blur}}
\hfil
\subfloat{\includegraphics[width=0.32\textwidth]{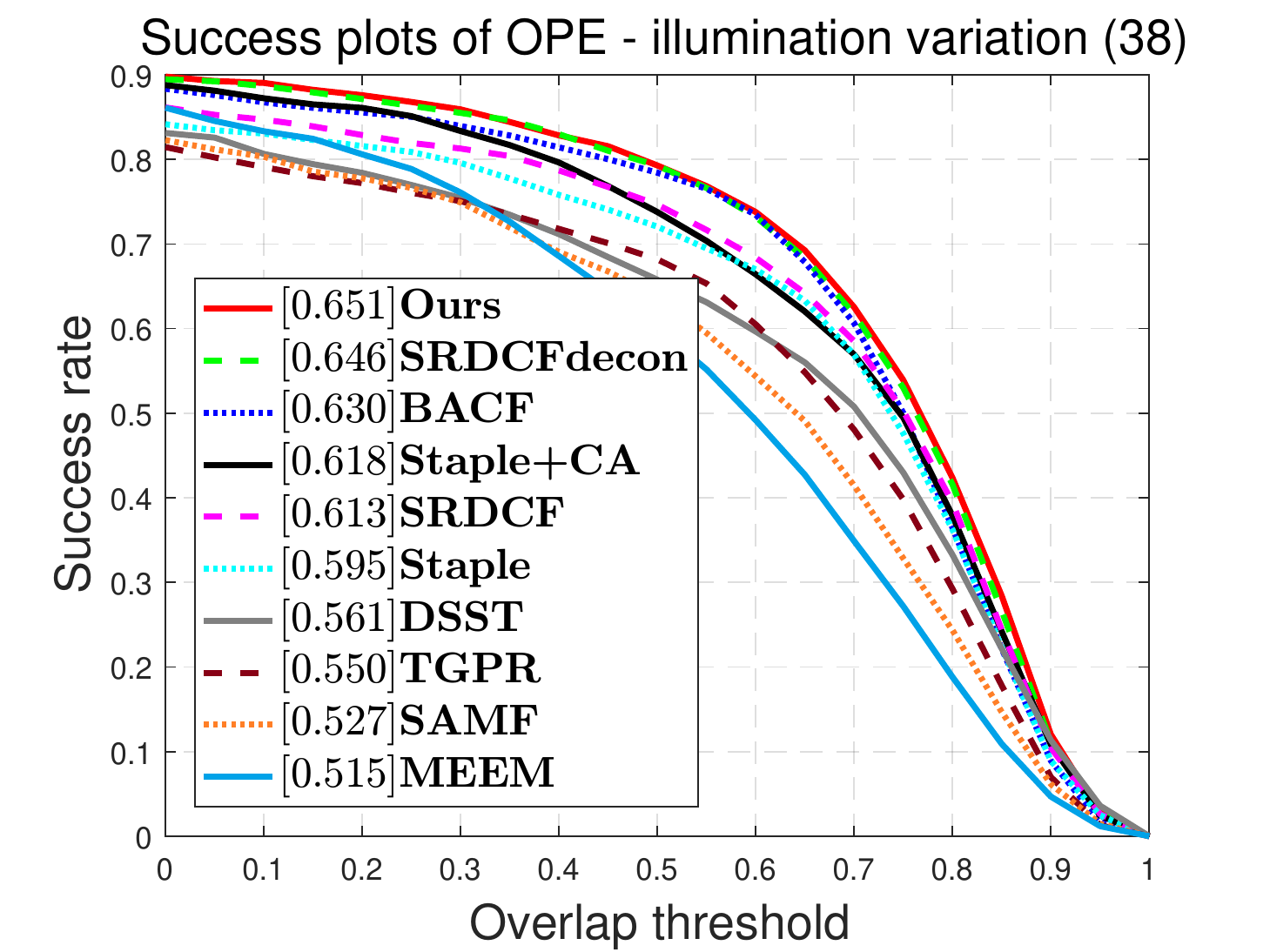}
\label{auc_occlusion}}
\hfil
\subfloat{\includegraphics[width=0.32\textwidth]{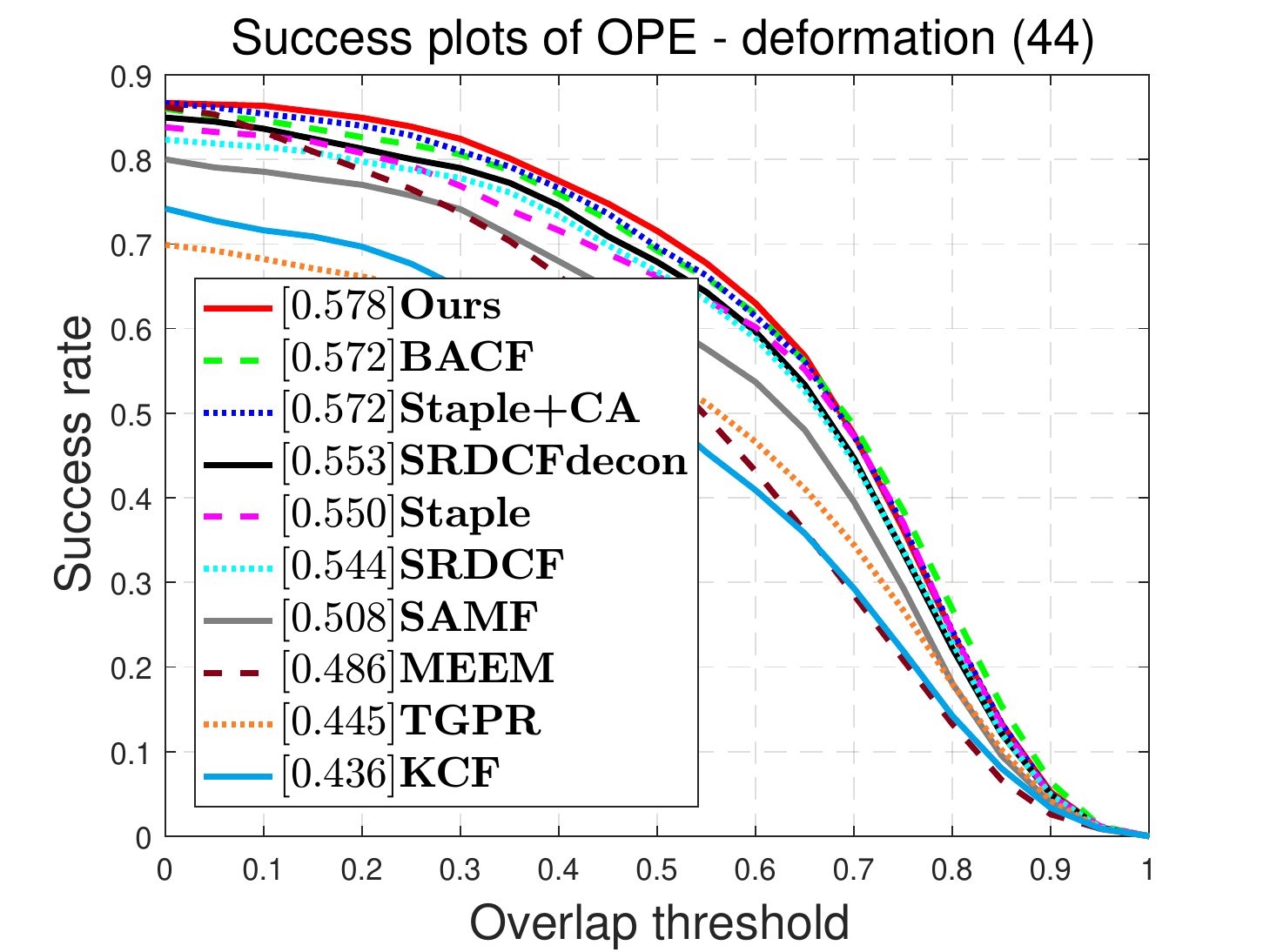}
\label{auc_deformation}}
\hfil
~
\centering
\subfloat{\includegraphics[width=0.32\textwidth]{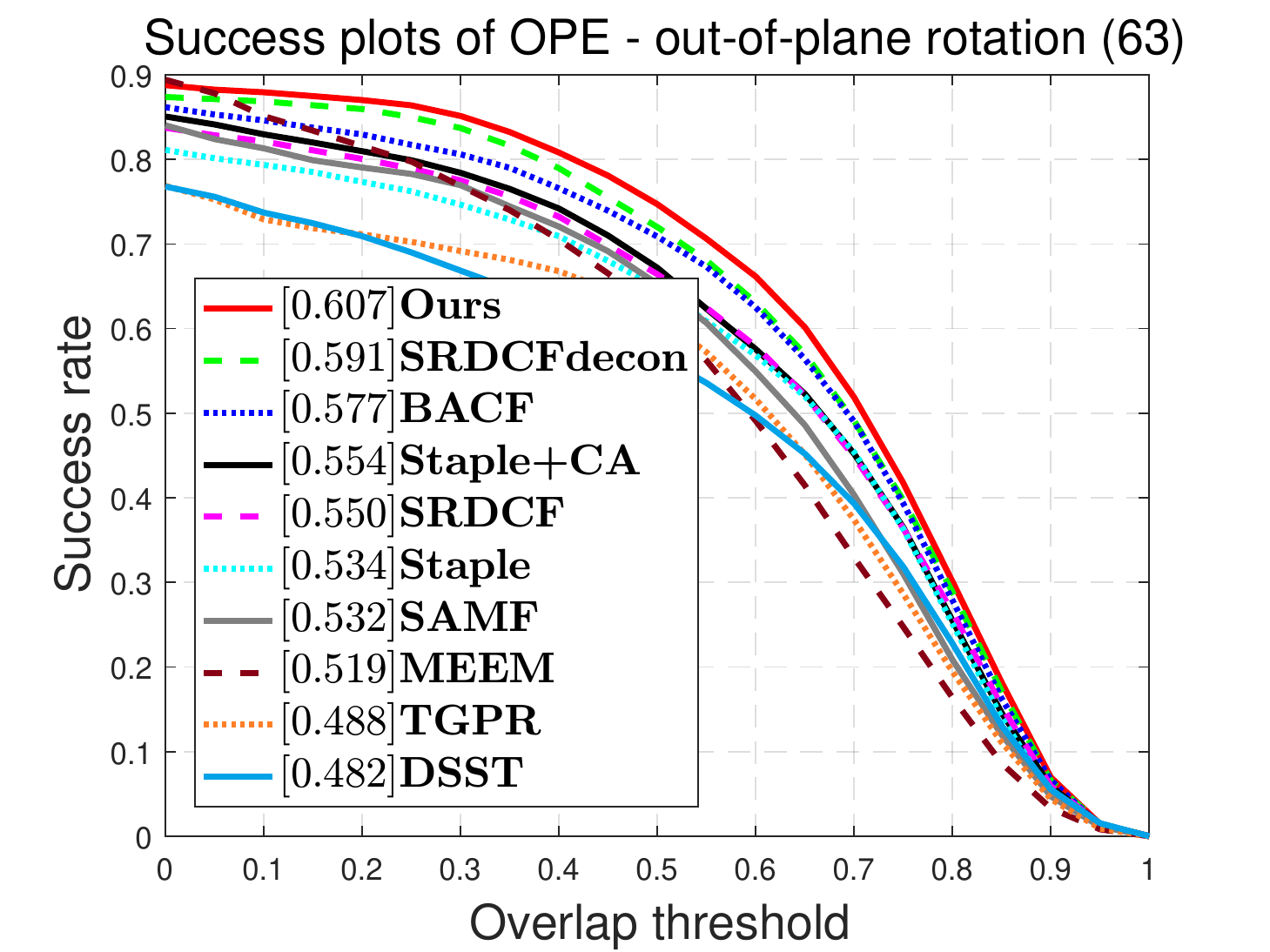}
\label{auc_background_clutter}}
\hfil
\subfloat{\includegraphics[width=0.32\textwidth]{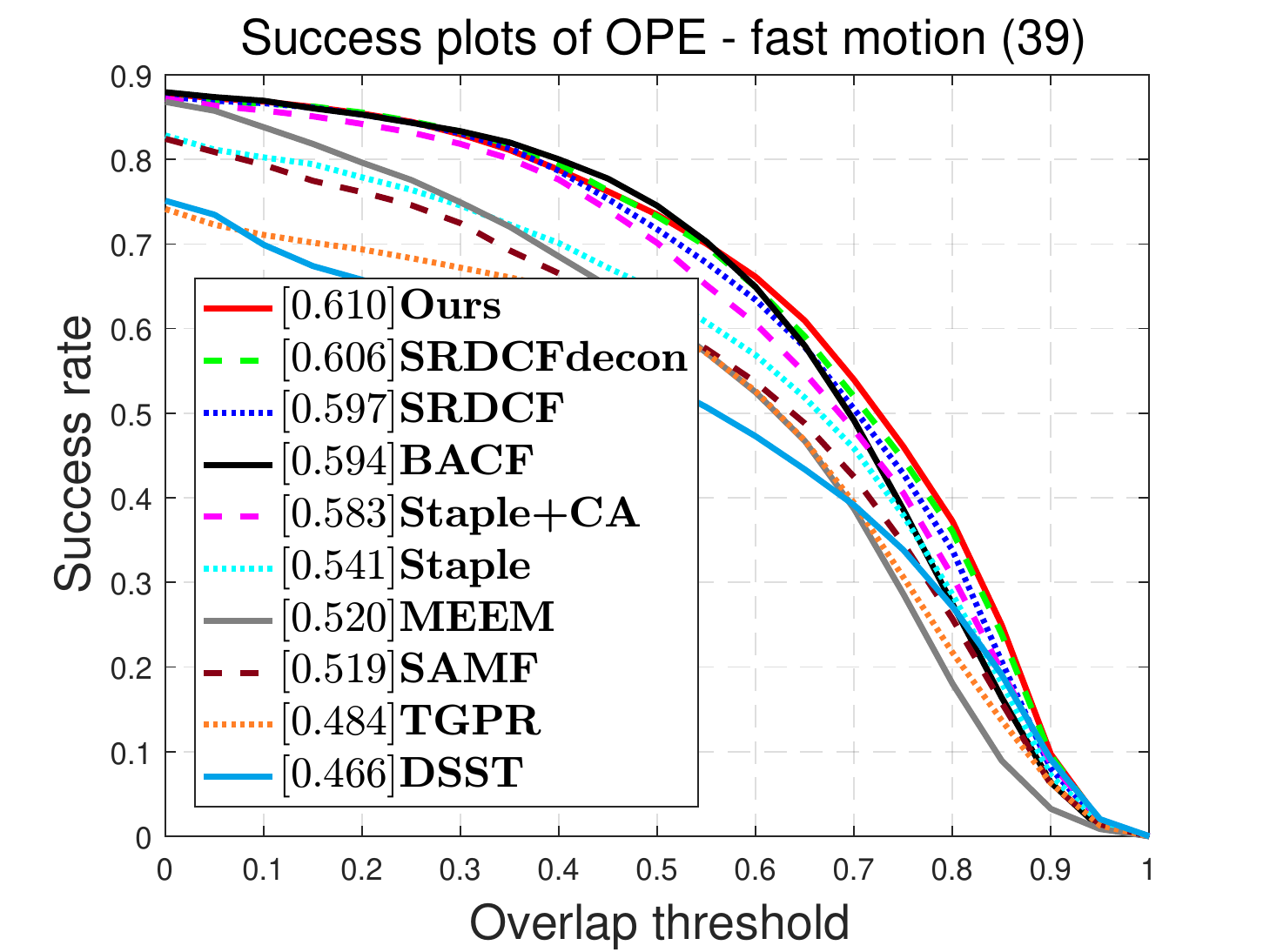}
\label{auc_scale_variation}}
\hfil
\subfloat{\includegraphics[width=0.32\textwidth]{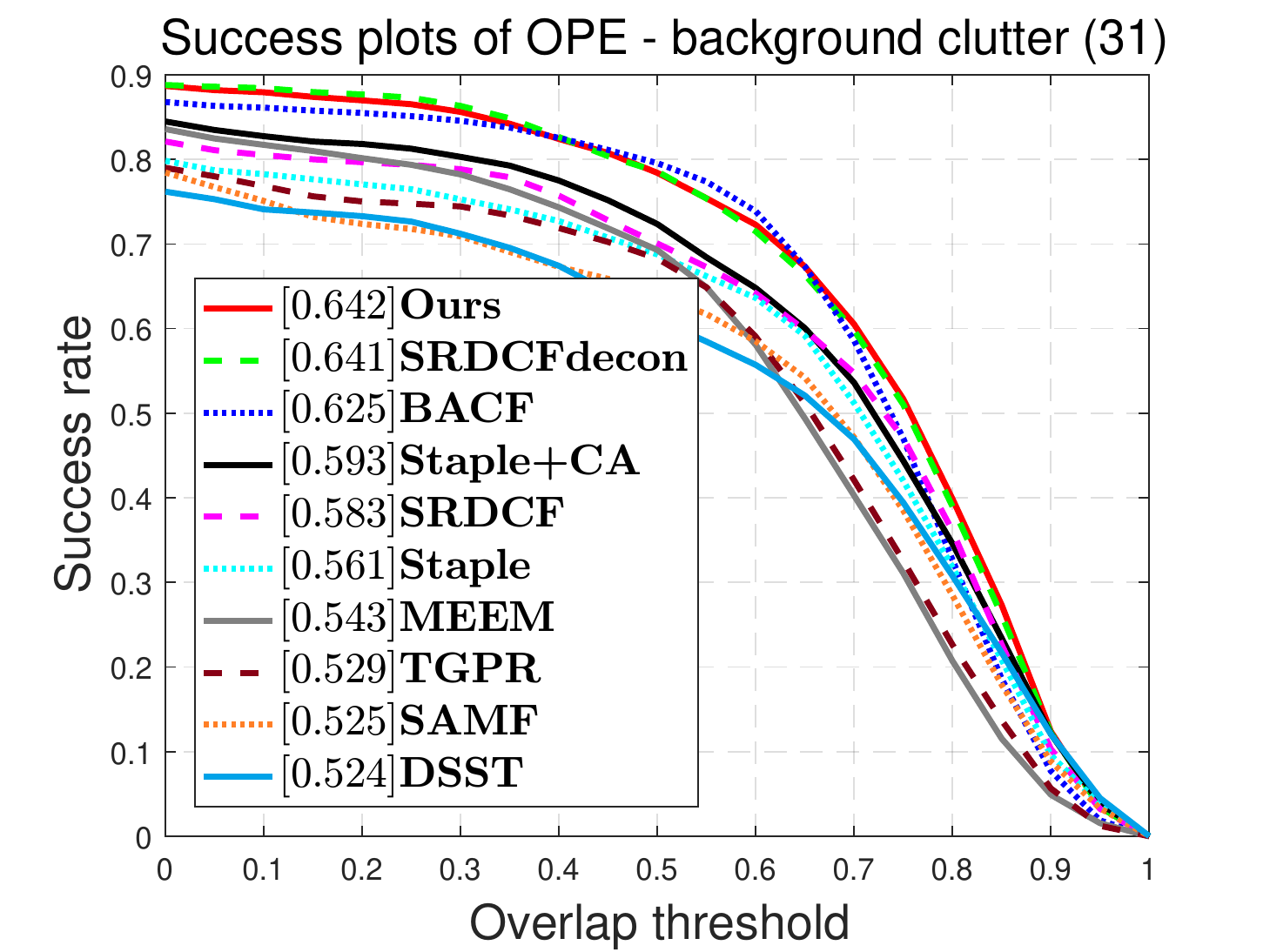}
\label{auc_outofplane_rotation}}
\caption{Attribute-based analysis of our approach on the OTB-2015 dataset. Success plots are shown for nine challenging attributes. Our proposed approach performs best against the state-of-art trackers.}
\label{fig:attrcomp}
\end{figure*}

\subsubsection{Attribute Based Comparison}
\begin{figure*}[!htbp]
\centering
\subfloat{\includegraphics[width=0.32\textwidth]{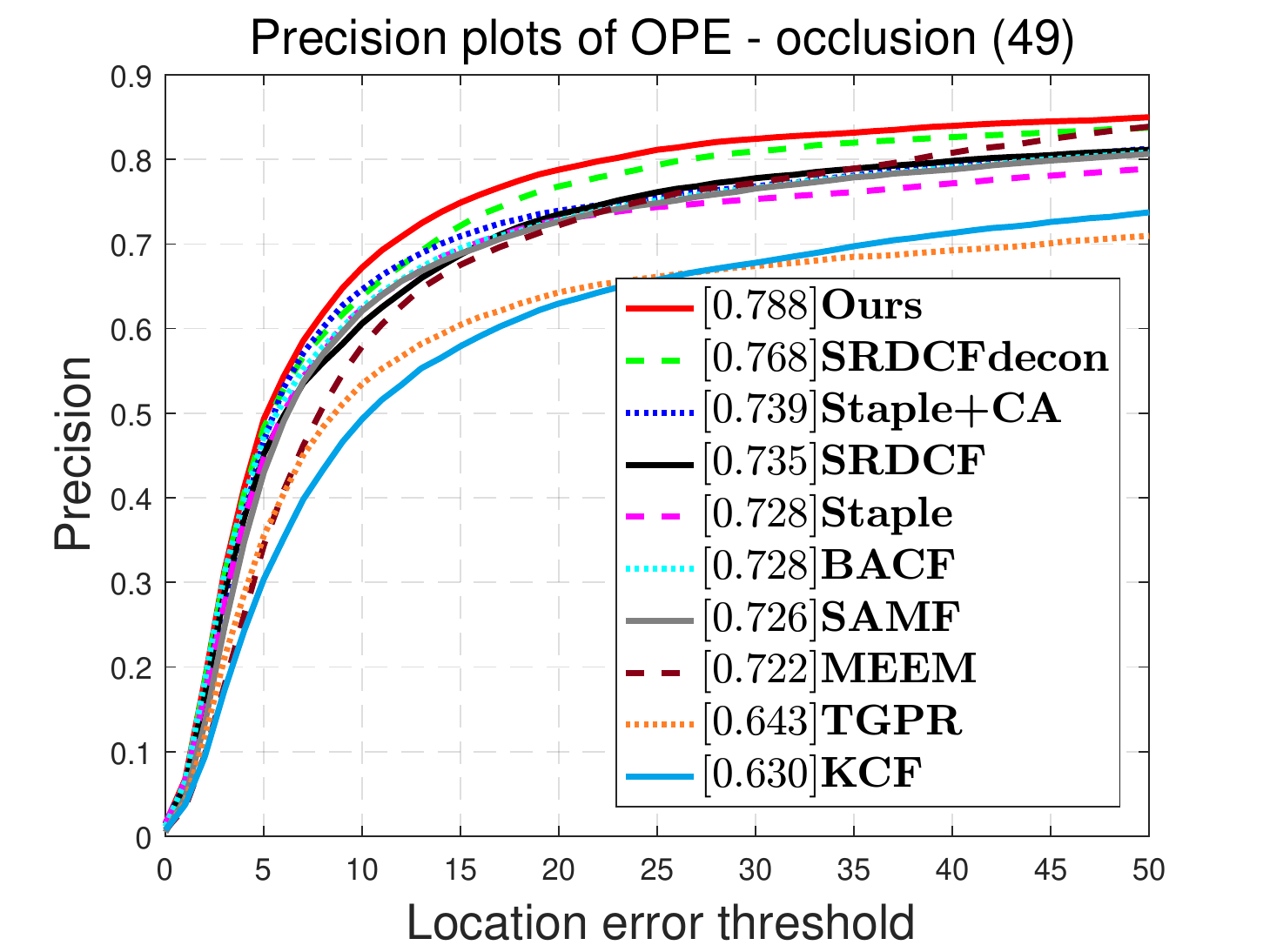}
\label{ps_fast_motion}}
\hfil
\subfloat{\includegraphics[width=0.32\textwidth]{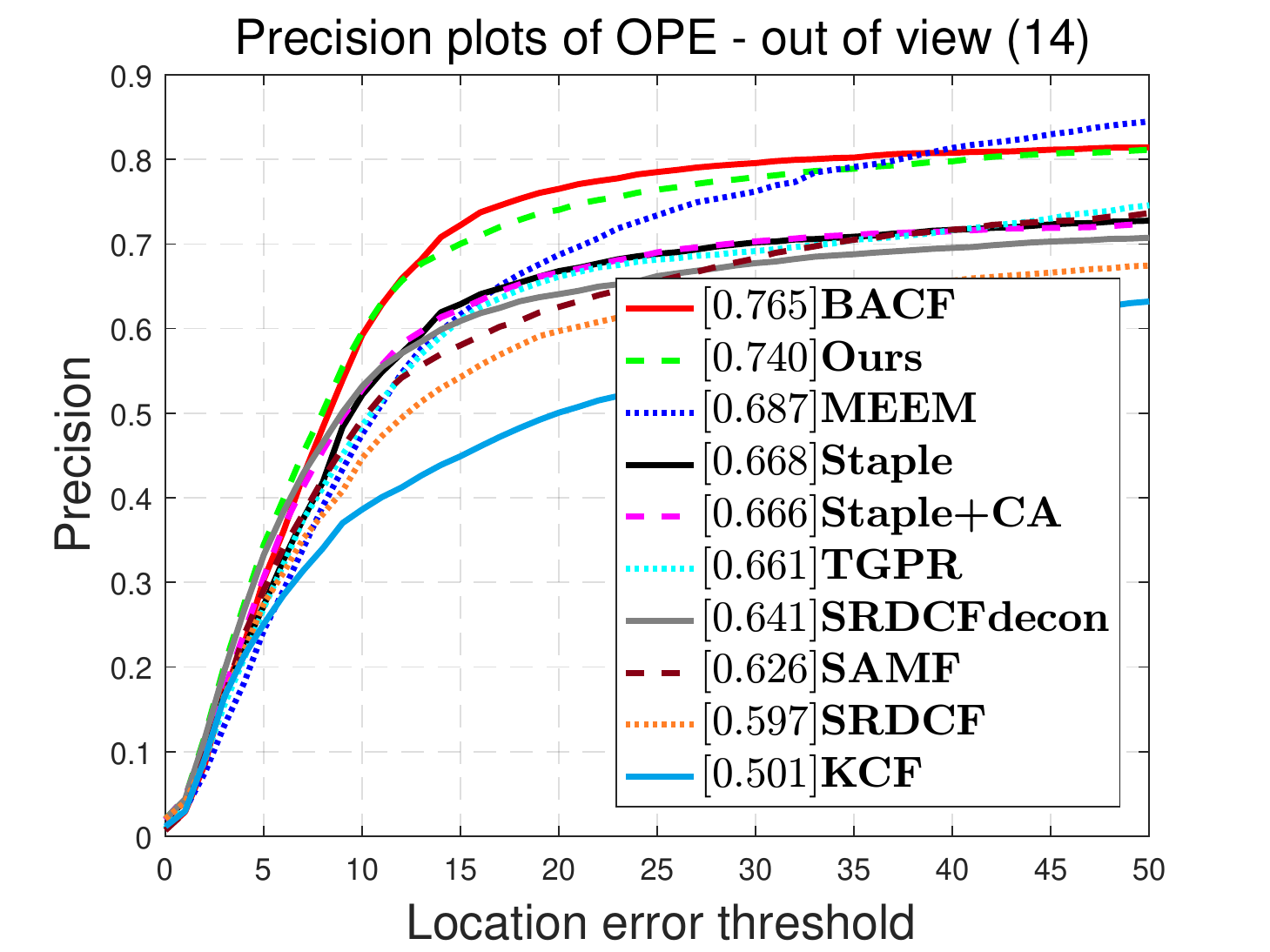}
\label{ps_outofview}}
\hfil
\subfloat{\includegraphics[width=0.32\textwidth]{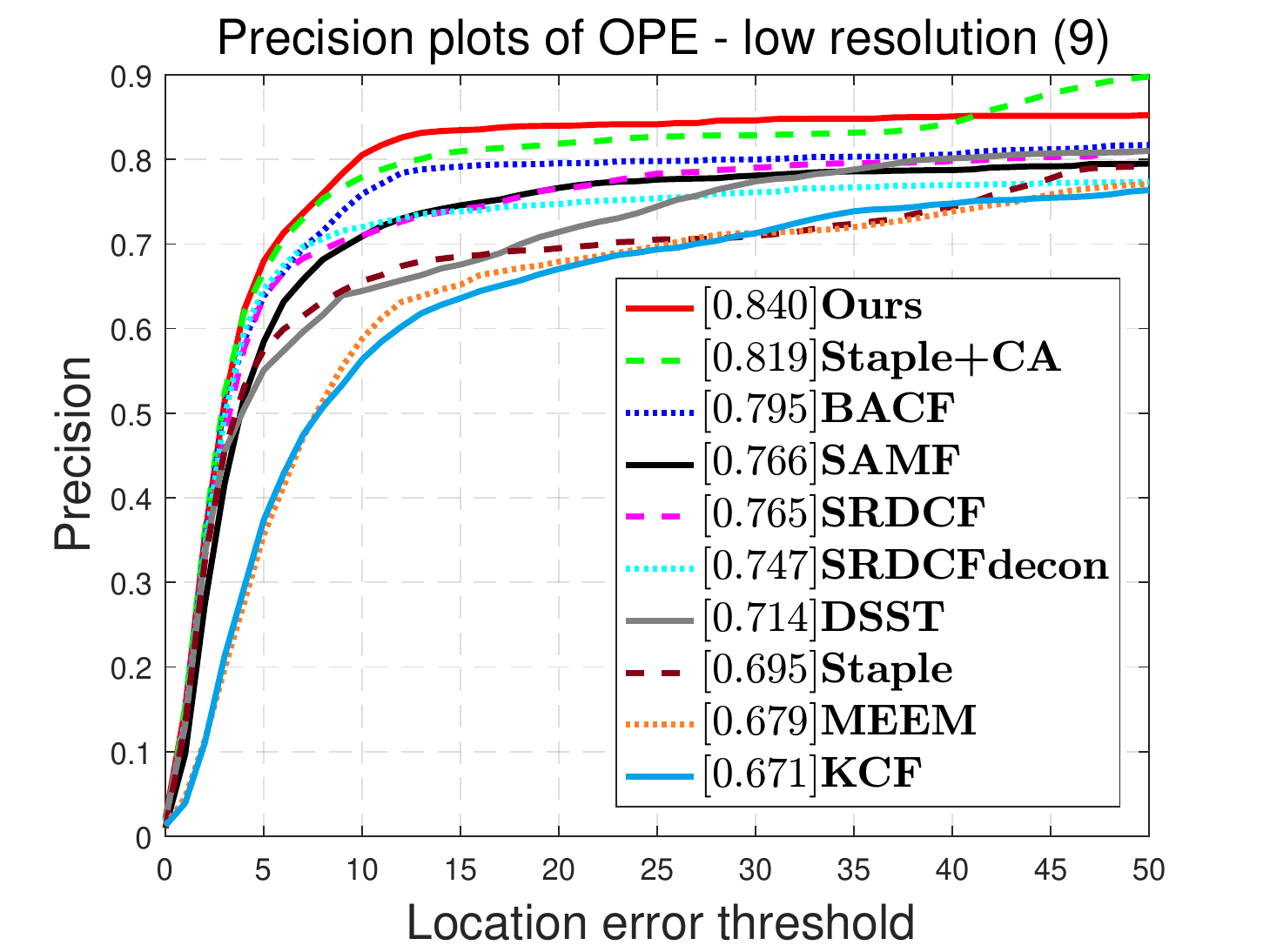}
\label{ps_low_resolution}}
\hfil
~
\centering
\subfloat{\includegraphics[width=0.32\textwidth]{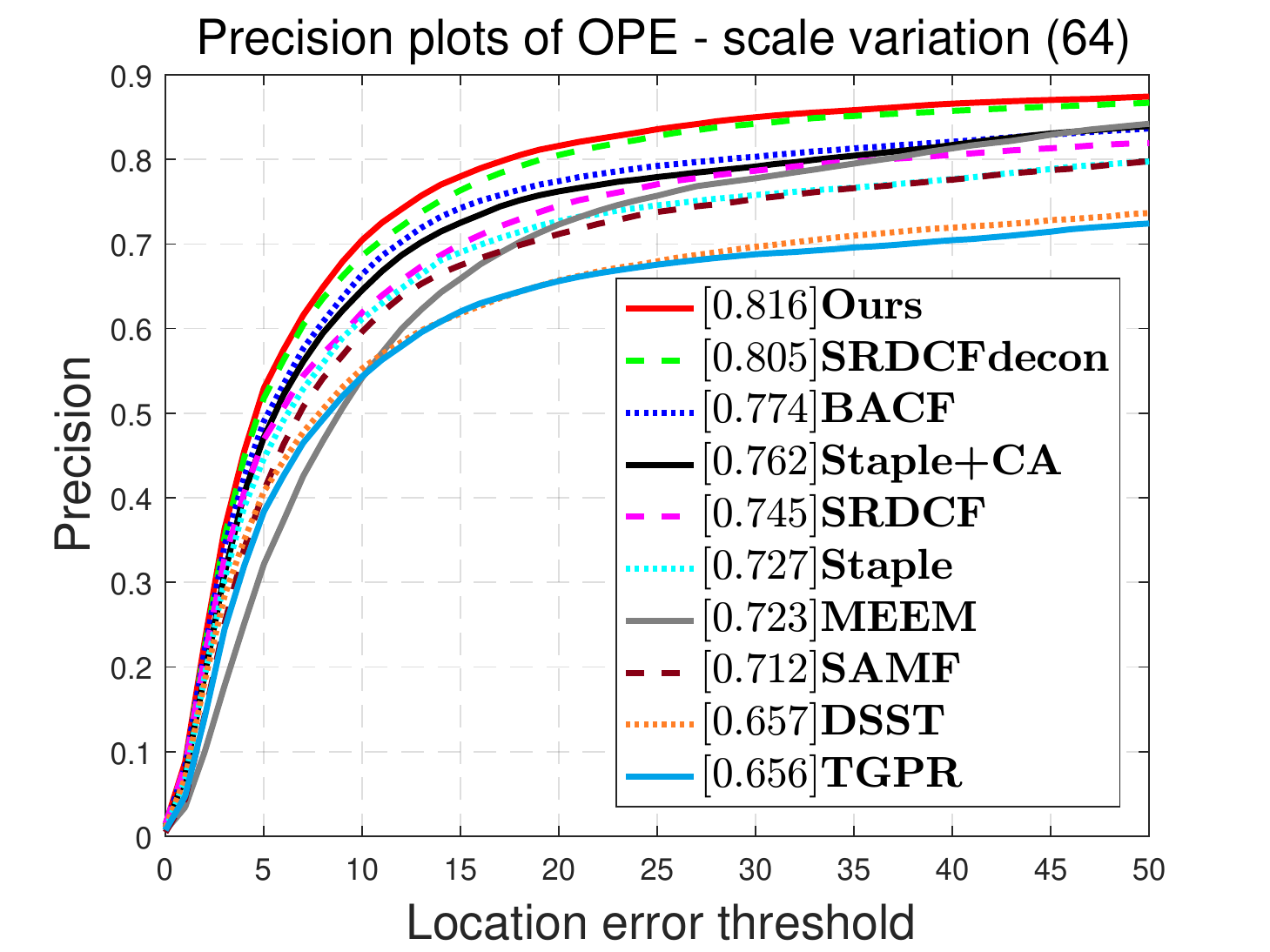}
\label{ps_motion_blur}}
\hfil
\subfloat{\includegraphics[width=0.32\textwidth]{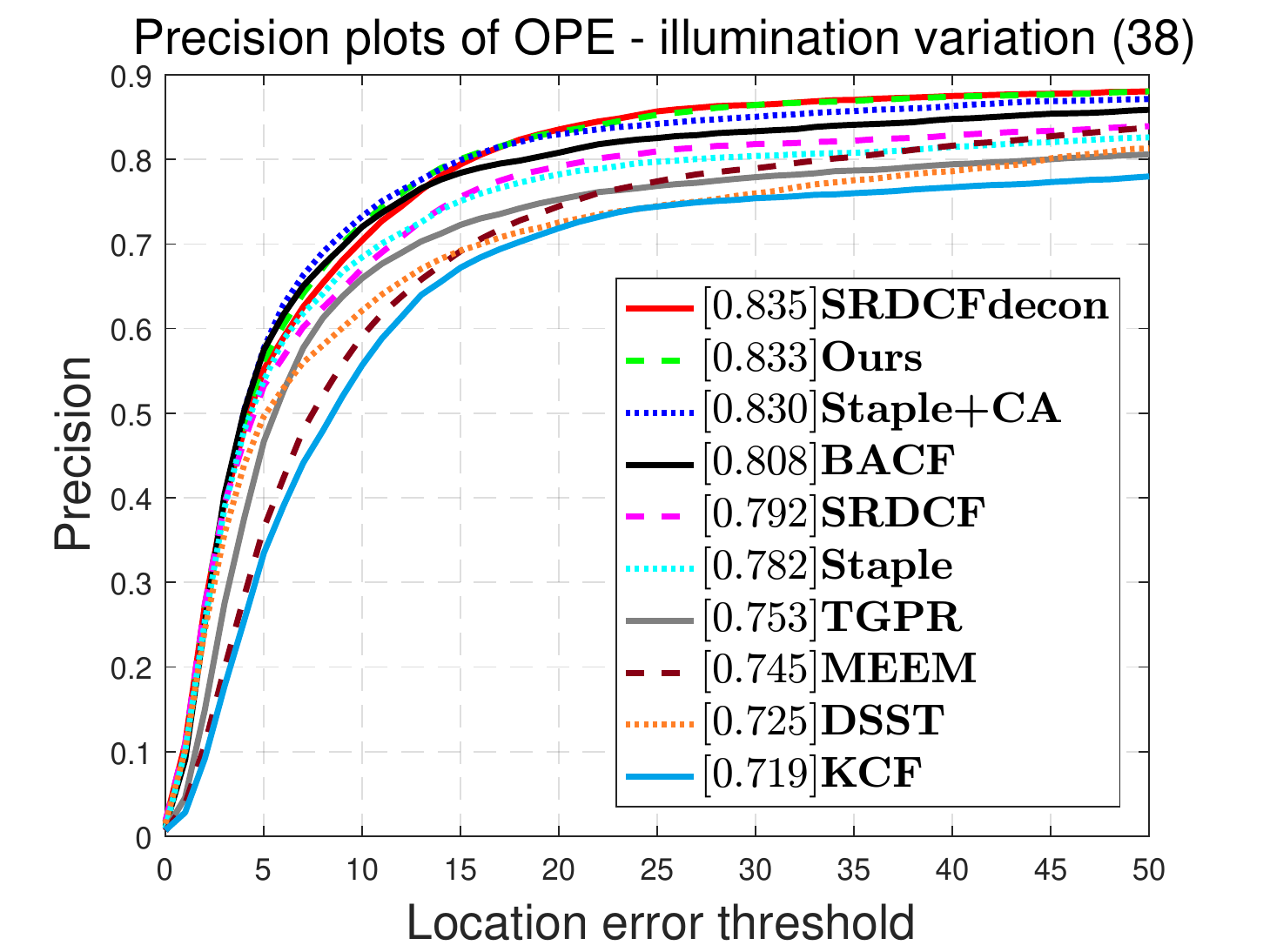}
\label{ps_occlusion}}
\hfil
\subfloat{\includegraphics[width=0.32\textwidth]{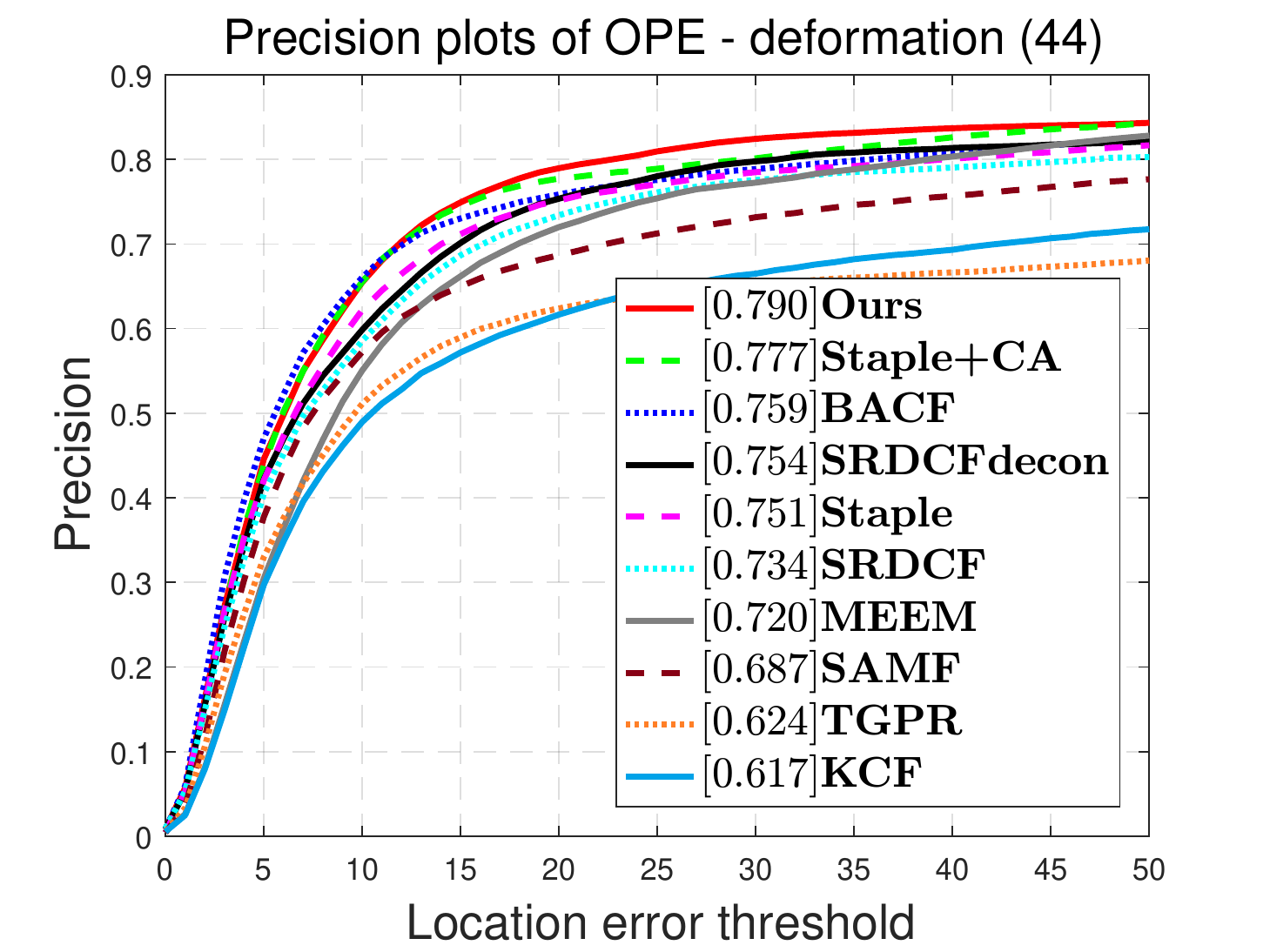}
\label{ps_deformation}}
\hfil
~
\centering
\subfloat{\includegraphics[width=0.32\textwidth]{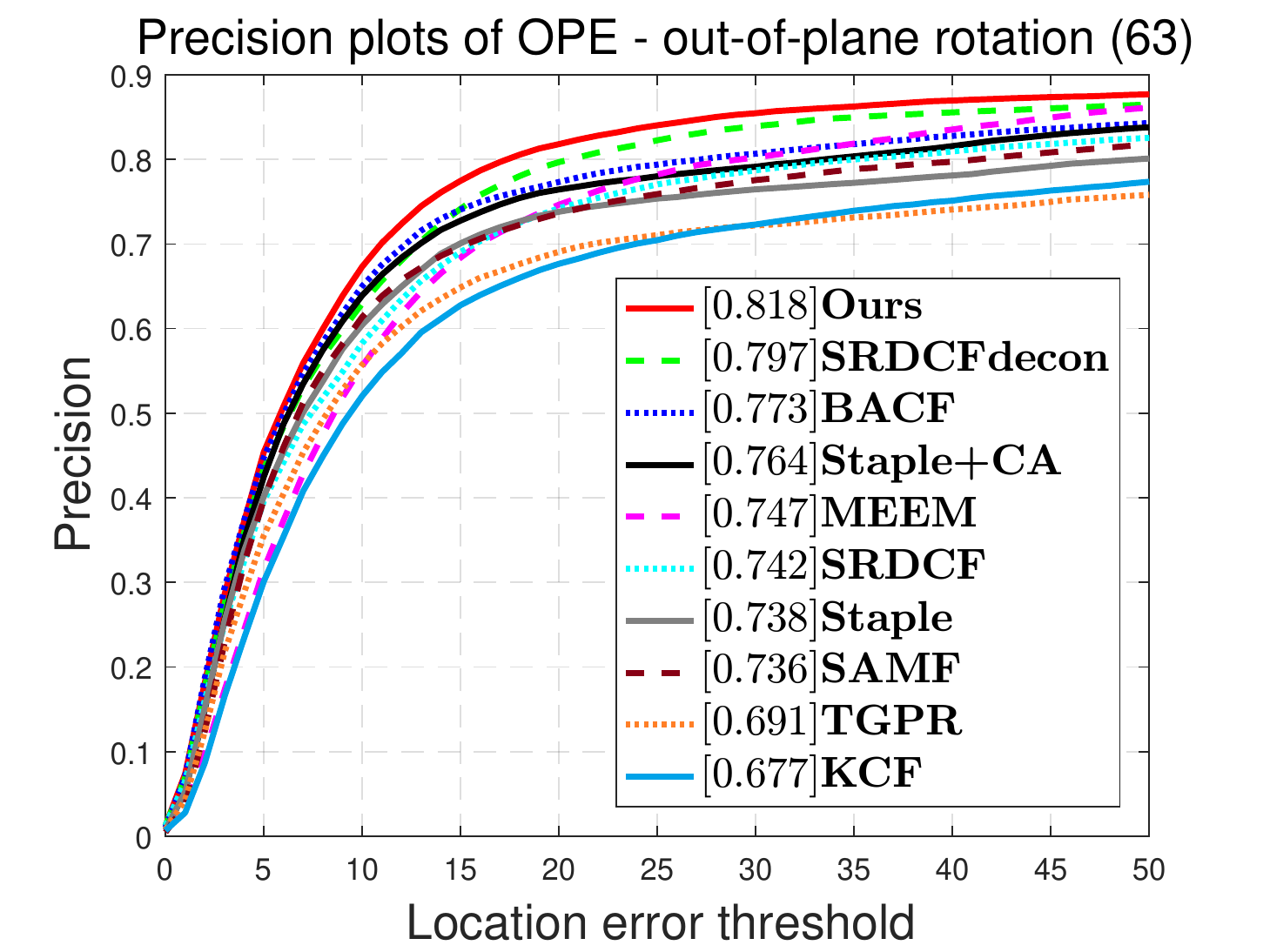}
\label{ps_background_clutter}}
\hfil
\subfloat{\includegraphics[width=0.32\textwidth]{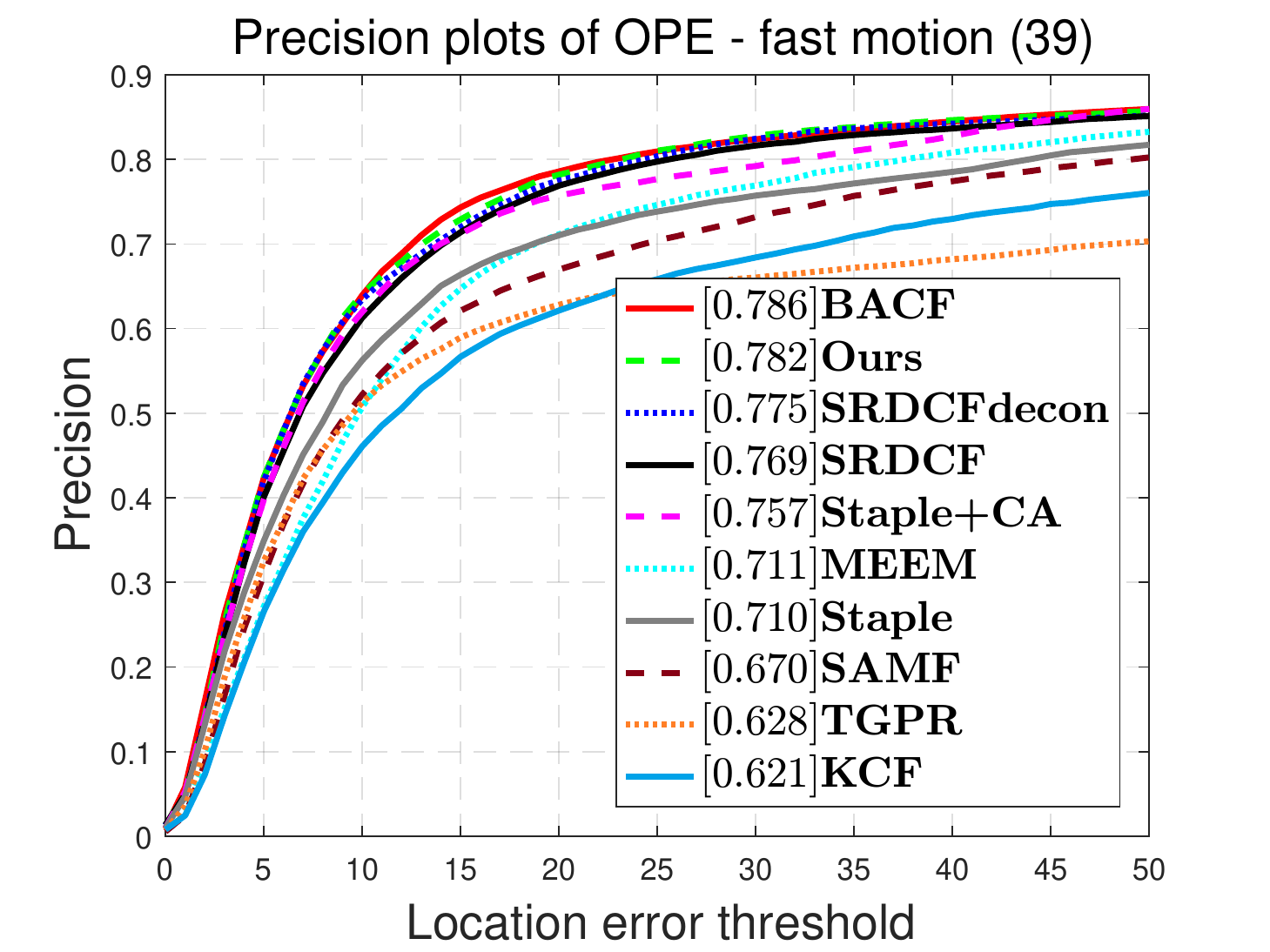}
\label{ps_scale_variation}}
\hfil
\subfloat{\includegraphics[width=0.32\textwidth]{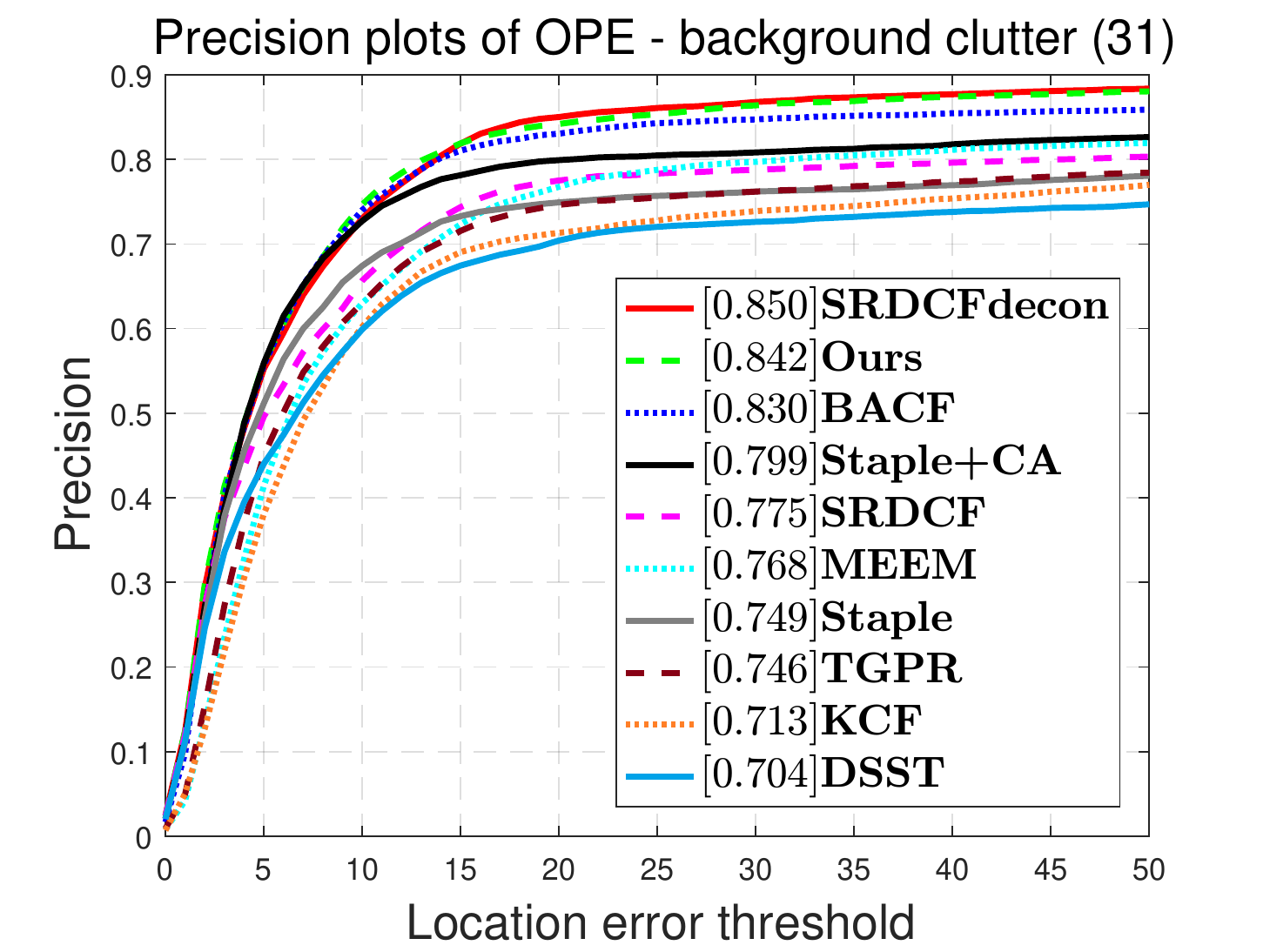}
\label{ps_outofplane_rotation}}
\caption{Attribute-based analysis of our approach on the OTB-2015 dataset. Precision plots are shown for nine challenging attributes. Our proposed approach is competitive with the state-of-art trackers.}
\label{fig:attrcomp2}
\end{figure*}

We evaluate our tracker by providing an attribute-based analysis on the OTB-2015 dataset. All the 100 sequences in the dataset are annotated with 11 different attributes, namely scale variation, occlusion, illumination variation, motion blur, deformation, fast motion, out-of-plane rotation, background clutters, out-of-view, in-plane rotation and low resolution. We only report the results of nine main challenging attributes in Figure~\ref{fig:attrcomp} and Figure~\ref{fig:attrcomp2}. For clarity, only the top 10 trackers in each plot are displayed. In case of occlusion, our approach achieves an AUC score of 60.0\% and a PS score of 78.8\%, providing a relative gain 2.9\% in term of AUC score and 7.2\% in term of PS score compared to the baseline SRDCF. In terms of AUC score, our proposed approach performs best against the state-of-art trackers. Overall, our tracking approach can achieve competitive results in all the attributes especially in occlusion, low resolution, scale variation, deformation and out-of-plane.

\begin{figure*}[!htbp]
\centering
\subfloat{\includegraphics[width=0.35\textwidth]{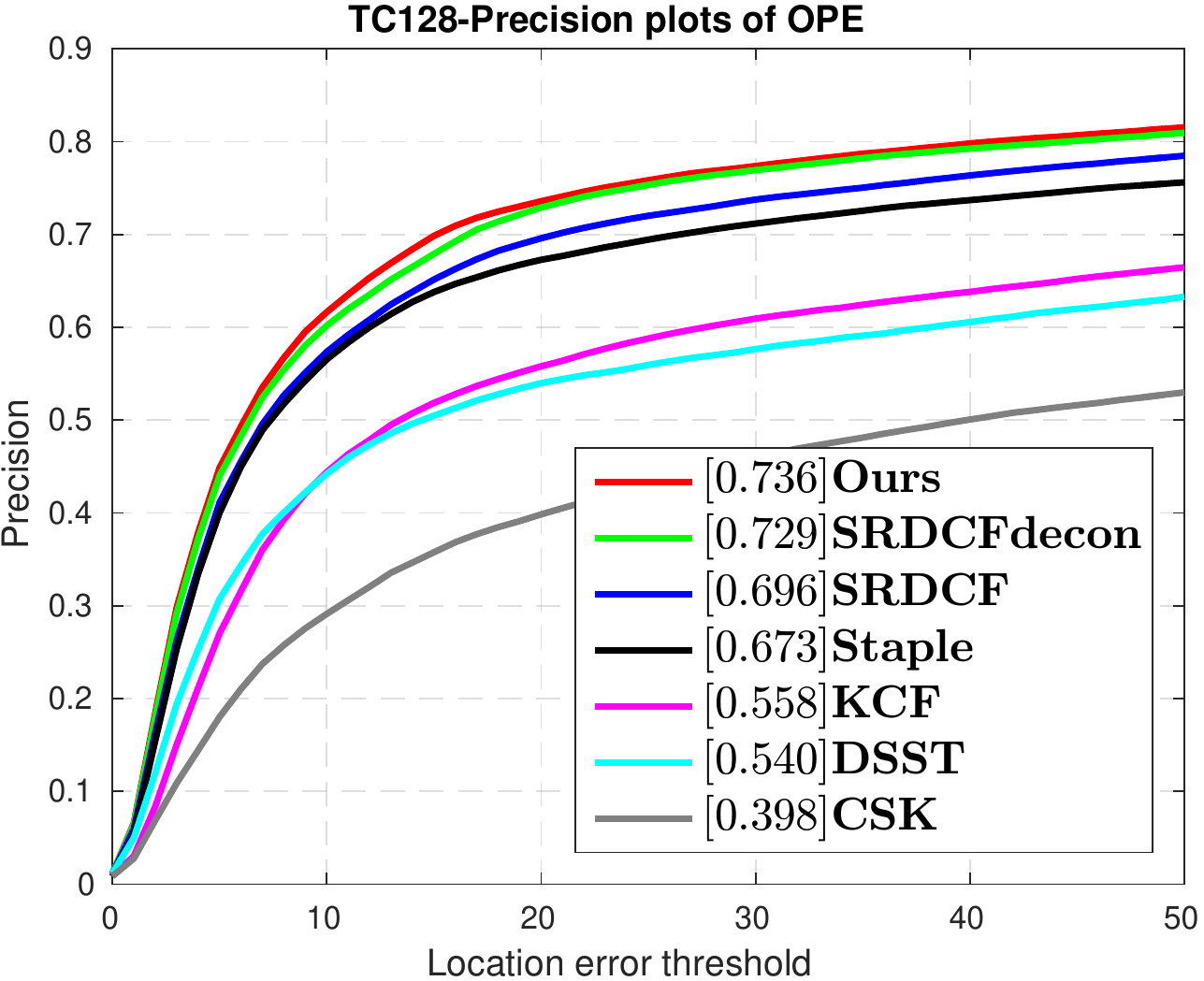}
\label{fig_tc128_precision}}
\hfil
\subfloat{\includegraphics[width=0.35\textwidth]{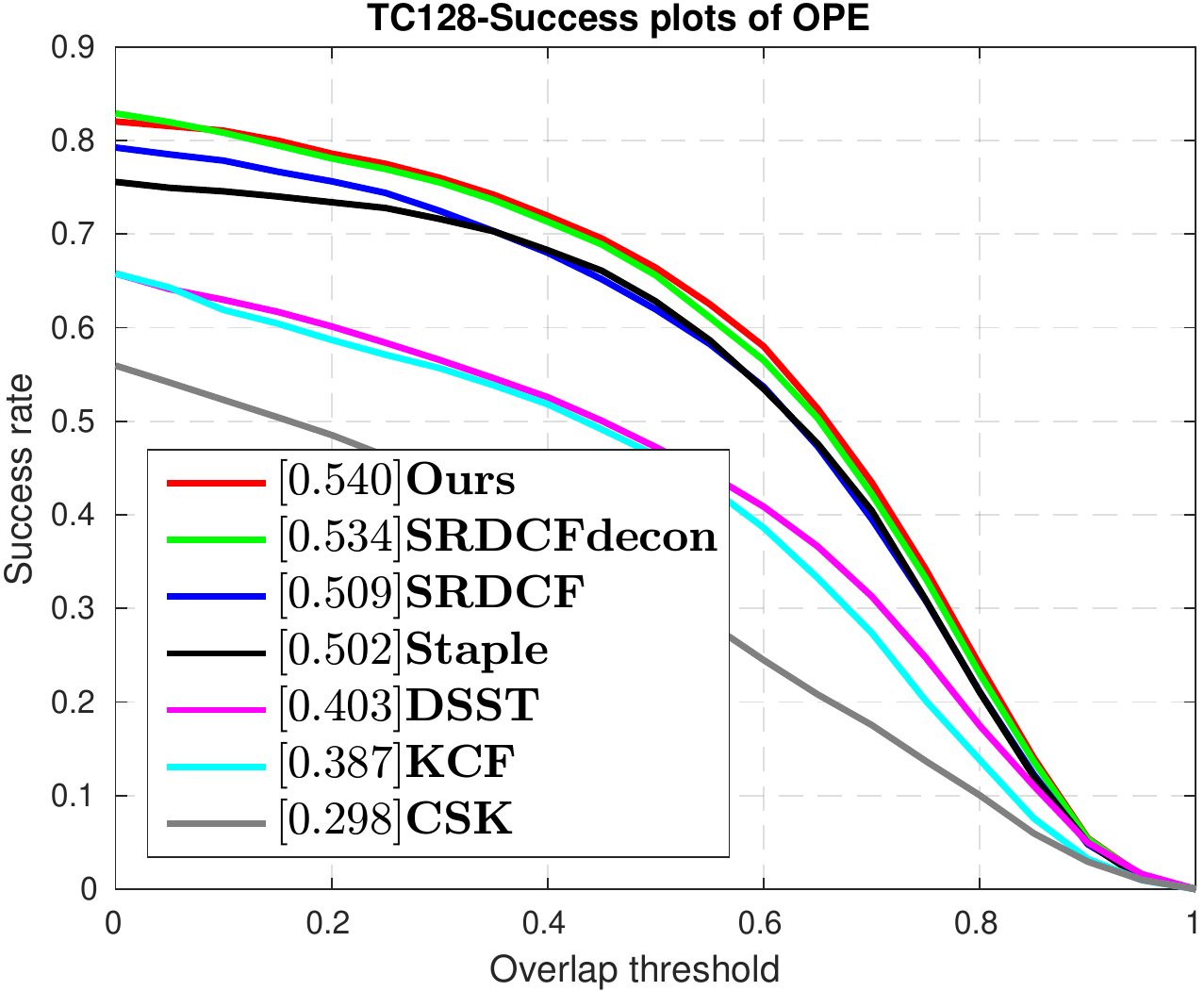}
\label{fig_tc128_success}}
\caption{Precision and success plots over the 128 sequences using one-pass evaluation on the Temple-color dataset. Our approach performs best against the state-of-art trackers.}
\label{fig:tc128results}
\end{figure*}
\subsubsection{Qualitative Results}
\begin{figure*}[!htbp]
\centering
\includegraphics[width=0.8\textwidth]{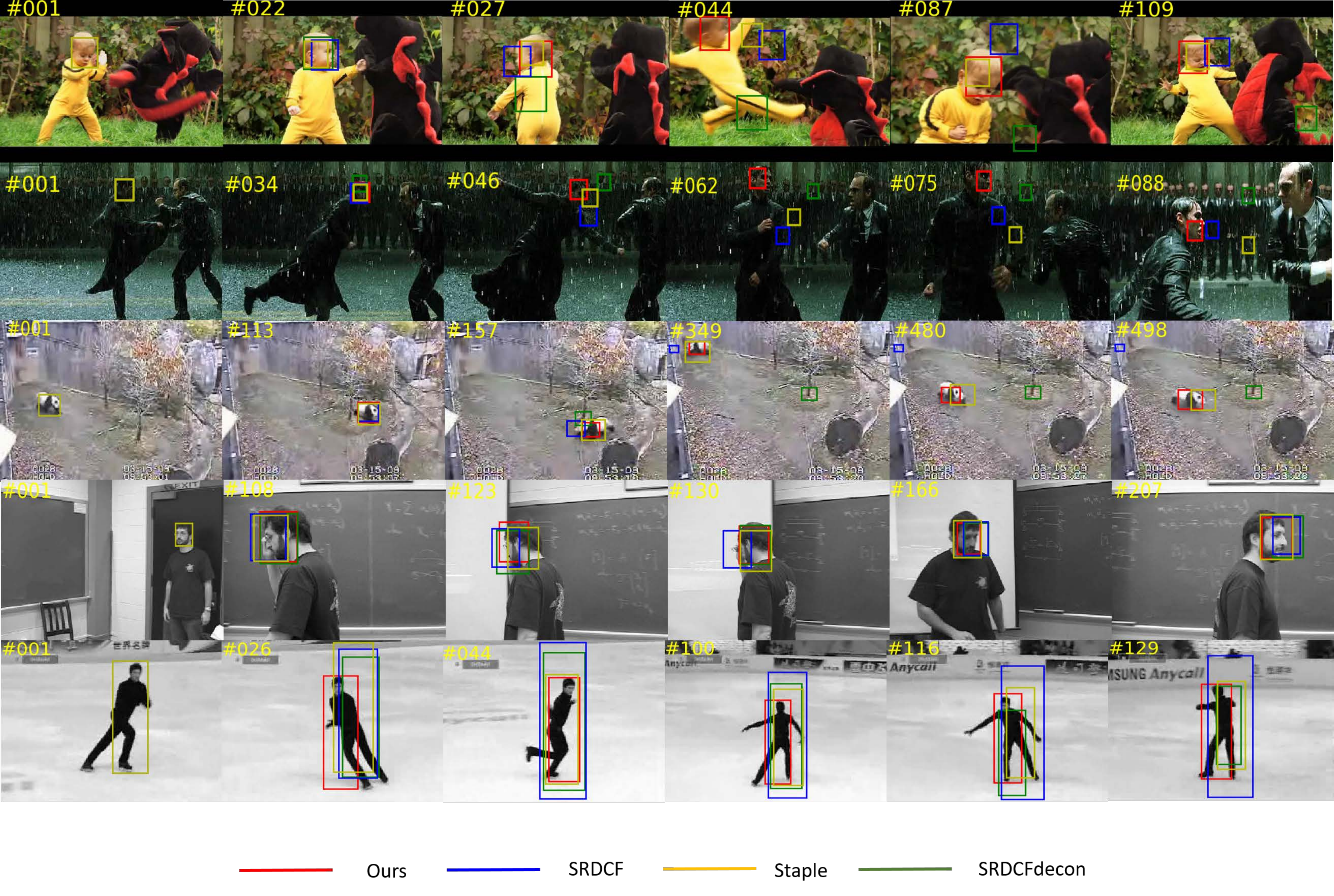}
\caption{A qualitative comparison of our proposed tracker with the state-of-art methods (Staple~\cite{Staple}, SRDCF~\cite{SRDCF} and SRDCFdecon~\cite{SRDCFdecon}) on the \textit{Dragonbaby}, \textit{Matrix}, \textit{Panda}, \textit{Freeman1}, and \textit{Skater2} sequences. The proposed approach performs favorably against these trackers.}
\label{fig:quares}
\end{figure*}

We compare our approach with the state-of-art methods (Staple~\cite{Staple}, SRDCF~\cite{SRDCF} and SRDCFdecon~\cite{SRDCFdecon}) on the \textit{Dragonbaby}, \textit{Matrix}, \textit{Panda}, \textit{Freeman1}, and \textit{Skater2} sequences in Figure~\ref{fig:quares}. Our method can effectively handle the challenging situations such as fast motion, low resolution and deformation, and achieve more visually feasible results against these trackers.

\subsubsection{Comparison with deep features-based trackers}
To further assess our tracker, we conduct experiments to compare our method with recently deep features-based trackers. We choose the deep version of the baseline tracker DeepSRDCF as our new baseline. The experimental results are shown in Fig.~\ref{fig:extra_hog}. It can be seen that our tracker outperforms the baseline DeepSRDCF by a large gain of 3.8\%. Our tracker can achieve similar performance with C-COT and MDNet and it is very competitive with ECO, MCCT and DRT. It is also worth noting that our tracker just uses first layer deep features while ECO, MCCT and DRT use either continuous operators or ensemble methods to fuse multiple deep layer features with different resolutions for boosting the performance.

\subsection{Temple Color Dataset}
Besides, we evaluate our proposed approach on the Temple-color dataset~\cite{TempleColor} with the state-of-art tracking methods, including CSK~\cite{CSK}, KCF~\cite{HenriquesC0B15}, DSST~\cite{DSST}, Staple~\cite{Staple}, SRDCF~\cite{SRDCF}, SRDCFdecon~\cite{SRDCFdecon}.The result in Figure~\ref{fig:tc128results} shows that our approach performs best against the state-of-art methods. Among the evaluated methods, SRDCF and SRDCFdecon respectively obtain a mean AUC of 50.9\% and 53.4\%, while our approach achieves an AUC score of 54.0\%.

\begin{table}[!htbp]
\caption{Baseline comparison on the VOT-2015 dataset, VOT-2016 dataset and VOT-2017 dataset.}
\label{tb:vot_baseline}
\begin{center}
\begin{tabular}{|c|l|c|c|c|}
\hline
dataset & trackers   & Accuracy & Robustness & EAO \\
\hline
\multirow{2}{*}{VOT-2015}
&SRDCF  &0.56  &  1.24   & 0.288  \\
\cline{2-5}
&\textbf{Ours}  & 0.57 &  1.08  &  0.309 \\
\hline
\multirow{2}{*}{VOT-2016}
&SRDCF  &0.52  &  1.50   & 0.247  \\
\cline{2-5}
&\textbf{Ours}  & 0.53 &  1.37  &  0.272 \\
\hline
\multirow{2}{*}{VOT-2017}
&SRDCF  &0.47  &  3.47   & 0.119  \\
\cline{2-5}
&\textbf{Ours}  & 0.50 &  2.91  &  0.151 \\
\hline
\end{tabular}
\end{center}
\end{table}

\subsection{VOT Dataset}
Finally, we present the baseline comparison on VOT-2015, VOT-2016 and VOT-2017 dataset, which all containing 60 challenging videos. The trackers are evaluated in terms of accuracy, robustness, and expected average overlap (EAO). The accuracy is the average overlap between the predicted and ground truth bounding boxes. The robustness measures how many times the tracker fails during the tracking. The EAO estimates the average overlap a tracker is expected to attain on a large collection of short-term sequences with the same visual property as the given dataset. Table~\ref{tb:vot_baseline} shows the baseline comparison results. It can be seen that our tracker performs better than the baseline SRDCF on VOT-2015, VOT-2016 and VOT-2017 by a gain of 2.1\%, 2.5\% and 3.2\% in EAO metric, respectively. The results demonstrate the effectiveness of our learning strategy.

\section{Conclusion}
In this paper, we have proposed a generic tracking framework based on
the progressive multi-stage optimization strategy for easy-to-hard sample selection
with a novel time-weighted and detection-guided
self-paced regularizer. The proposed tracking scheme has the capability of
progressively expanding the reliable sample volume involved in discriminative learning
and capturing the multi-grained local distribution structure information of object samples,
resulting in tolerating relatively large intra-class variations while maintaining inter-class separability.
Moreover, we jointly optimize the self-paced learning strategy in conjunction with
the discriminative tracking process, leading to robust tracking results.
Experimental results over the benchmark video sequences have justified the effectiveness
and robustness of the proposed tracking framework.

\appendix
Here, we mainly present the detailed derivations of the solution for the sample weights $v^{n}$ in Eq. (12) (the derivations for Eq. (5), and Eq. (8) are similar). An each learning stage $n$ in Eq. (12), when we fix the parameters $\theta_{n}$ of the appearance model, minimizing the optimization problem Eq. (11) with respect to the sample weights $v^{n}$ is then equivalent to solving the following optimization problem,
\begin{equation}
\hspace{-2em}\min_{v^n\in[0,1]^{t}} \varepsilon(v^n) = \sum_{k=1}^{t}v_{k}^nl_{k}+\lambda_n \sum_{k=1}^{t}(\frac{1}{2} \frac{({v_{k}^{n}})^2}{\rho_{k}} - v_{k}^n ) + \xi \sum_{k=1}^{t}c_{k}v_{k}^n.
\end{equation}
The $\varepsilon(v^n)$ is a convex quadratic function with respect to the sample weights $v^{n}$. The partial derivative $\varepsilon(v^n)$ with respect to $v_{k}^{n}$ gives,
\begin{equation}
  \frac{\partial \varepsilon(v^n)}{\partial v_{k}^{n}} = l_{k} + \lambda_n \frac{v_{k}^{n}}{\rho_{k}} - \lambda_n + \xi c_{k}
\end{equation}
The stationary point is calculated by setting the partial derivative to zero,
\begin{equation}
    \frac{\partial \varepsilon(v^n)}{\partial v_{k}^{n}} = 0 \Longleftrightarrow v_{k}^{n} = \rho_{k} - \frac{l_{k} + \xi c_{k}}{\lambda_n}\rho_{k}
\end{equation}
Considering the condition $v_{k}^n\in[0,1]$, the closed-form optimal solution for $v_{k}^{n}$ can be derived as (corresponding to Eq. (12)):
\begin{equation}
  v_{k}^{n} = \left\{
  \begin{array}{lcl}
    \rho_{k} - \frac{l_{k} + \xi c_{k}}{\lambda_n}\rho_{k}, && l_{k} + \xi c_{k}< \lambda_n; \\
    0, && l_{k} + \xi c_{k} \geq \lambda_n.
  \end{array}\right.
\end{equation}


%





\ifCLASSOPTIONcaptionsoff
  \newpage
\fi


\normalem
\bibliographystyle{IEEEtran}
\bibliography{splt}

\end{document}